 \documentclass[alpha-refs]{wiley-article}

\usepackage{siunitx}
\usepackage{anyfontsize}
\usepackage{lmodern}
\usepackage{caption}
\usepackage{multirow}
\usepackage{footnote}
\usepackage{hyperref}
\usepackage{amsmath}
\usepackage[ruled]{algorithm2e}
\usepackage{algpseudocode}
\usepackage{longtable}
\usepackage{amssymb}
\usepackage{booktabs}
\DeclareMathOperator*{\argmax}{Arg\,Max}
\papertype{Research Article}

\title{A Vision-Based Navigation System for Arable Fields}

\author[1]{Rajitha de Silva}
\author[1]{Grzegorz Cielniak}
\author[1]{Junfeng Gao}

\affil[1]{Lincoln Agri-Robotics Centre, Lincoln Institute for Agri-Food Technology, University of Lincoln, Lincoln, United Kingdom}

\corraddress{Junfeng Gao, Lincoln Agri-Robotics Centre, Lincoln Institute for Agri-Food Technology, University of Lincoln, Lincoln, United Kingdom}
\corremail{jugao@lincoln.ac.uk}

\fundinginfo{This work was supported by Lincoln Agri-Robotics as part of the Expanding Excellence in England (E3) fund by UKRI's Research England.}

\runningauthor{Rajitha de Silva et al.}

\begin{document}

\begin{frontmatter}
\maketitle

\begin{abstract}
Vision-based navigation systems in arable fields are an underexplored area in agricultural robot navigation. Vision systems deployed in arable fields face challenges such as fluctuating weed density, varying illumination levels, growth stages and crop row irregularities. Current solutions are often crop-specific and aimed to address limited individual conditions such as illumination or weed density. Moreover, the scarcity of comprehensive datasets hinders the development of generalised machine learning systems for navigating these fields. This paper proposes a suite of deep learning-based perception algorithms using affordable vision sensors for vision-based navigation in arable fields. Initially, a comprehensive dataset that captures the intricacies of multiple crop seasons, various crop types, and a range of field variations was compiled. Next, this study delves into the creation of robust infield perception algorithms capable of accurately detecting crop rows under diverse conditions such as different growth stages, weed density, and varying illumination. Further, it investigates the integration of crop row following with vision-based crop row switching for efficient field-scale navigation. The proposed infield navigation system was tested in commercial arable fields traversing a total distance of 4.5 km with average heading and cross-track errors of 1.24$^\circ$ and 3.32 cm respectively. 

\end{abstract}
\end{frontmatter}

\section{Introduction}

The global agricultural sector is at a critical juncture amidst the increasing food demand due rising global population, labour shortages, environmental sustainability and the drive for maximising the efficiency of food production with precision agricultural practices. Addressing the economic impacts of labour shortages and inefficient agricultural practices is simultaneously important while achieving sustainable food production minimising adverse environmental impact. Therefore, exploring technological solutions that mitigate these labour and economic challenges while remaining acutely conscious of their environmental impact, ensuring a harmonious balance between productivity and ecological preservation should be inherent in future agricultural robotic solutions~\citep{bogue2023robots}. The state-of-the-art (SOTA) solutions in agricultural robotic navigation void of these characteristics due to higher costs and poor reliability accommodates the need to develop cheaper vision-based solutions to fill in the demands of future agricultural automation. The integration of robotics and autonomous systems in agriculture has enabled precision farming operations leading to effective use of time and resources. 

Achieving autonomous navigation in agricultural robots is an enabling technology that must be optimised for the deployment of autonomous robots for precision agriculture. The existing autonomous navigation solutions for in-field navigation, albeit being efficient, often rely on expensive sensors such as Real-Time Kinematic Global Navigation Satellite System (RTK-GNSS) sensors. Camera-based agricultural robot navigation systems are a popular alternative to these expensive sensors~\citep{bonadies2019overview}. Implementation of such vision systems is often limited to crop row following behaviour~\citep{huang2020feedforward}. In most such systems, row switching is achieved by the aid of GNSS sensors or multiple cameras to identify row end and re-entry to the next row~\citep{bonadies2019overview, ahmadi2020visual, kanagasingham2020integrating}. Identification of the initial turning direction and the last crop row to be traversed while navigating an entire field is also important for this class of navigation algorithms to become self-reliant. Both of these problems entail the detection of the robot being at the crop row next to the edge of the field. The initial turn direction and the last row detection are often not discussed in the existing field scale navigation systems. Most existing methods require the user to define these parameters by declaring the initial turn direction and number of crop rows.

The premise of vision-based navigation in agricultural robots is to reduce the cost of the overall robotic system with the aim of increased adoption of such technologies. To this end, a vision-based navigation system that uses a single front-mounted camera to perform in-row navigation and row switching would be in line with this objective. In our previous work, we have developed a vision-based crop row detection algorithm~\citep{de2023deep} and an in-row navigation framework~\citep{de2022vision} that then uses the crop row detection to guide the robot through a single crop row only based on RGB images. The crop row switching algorithm presented in this paper relates to our previous work on vision-based crop row detection~\citep{de2023deep} and navigation~\citep{de2022vision} in arable fields. The newly proposed crop row switching algorithm serves as a vital bridge, seamlessly connecting with the established in-row navigation algorithm to dexterously integrate into a comprehensive, fully autonomous field-scale navigation behaviour. This row-switching algorithm could also integrated seamlessly with any other existing in-row navigation methods to eliminate the need for GNSS sensors in row-switching. This existing system can follow a crop row based on RGB image input and it can also identify the location of the end of crop rows when the robot is reaching towards the headland area. The crop row switching algorithm presented is be triggered upon the detection of the end of row (EOR) and navigates the robot towards the entry point of the next crop row to be traversed.

A complete field scale navigation system could be realised by facilitating the row-following behaviour with a complementary crop row switching algorithm to enable the headland traversal to switch between adjacent crop rows during infield navigation. Existing vision-based methods of crop row switching in arable fields require multiple cameras~\citep{xue2012variable} and symmetric robotic setups~\citep{ahmadi2021towards}. Certain existing approaches demand hybrid vision and GNSS solutions where row switching behaviour is dependent on the GNSS-based navigation~\citep{winterhalter2021localization}. The lack of a vision-only single camera-based solution hinders the cost-effective nature of vision-based navigation systems with the addition of multiple specialised camera setups and GNSS sensors. The initial row switching direction must be manually specified in the existing methods. The existence of autonomous field coverage direction identification is also an important feature of field scale navigation algorithms which is attributed to a user configuration parameter in the existing systems~\citep{bonadies2019overview}. Such improvements would be highly valued at the adoption of such agricultural robotic systems by the end users where the system functions fully autonomously without needing to configure to each specific deployment instance. We have developed a field-scale crop row navigation system by integrating our vision based in-row navigation methods~\citep{de2023deep, de2022vision} with the crop row switching and infield orientation algorithms introduced in this paper. 

\noindent The main contributions of this work are as follows:

\begin{itemize}
  \item A fully vision-based infield navigation system that can perform field scale navigation in crop row fields.
  \item A crop row switching pipeline based on vision and robot wheel odometry to navigate the robot towards the entry point of the next crop row to be traversed.
  \item An vision-based infield orientation algorithm that identifies the initial turning direction for the robot when reaching the end of first crop row in a field.
  \item A comprehensive evaluation and validation of the autonomous infield navigation system in simulation and a real arable field with a mobile robot deployed with the proposed system.
  
\end{itemize}

The remainder of this paper is arranged as follows: Section \ref{sec:bgr} discusses the existing approaches and their shortcomings. Section \ref{sec:vbaf} introduces the vision-based navigation paradigm in arable fields explaining the steps followed during infield navigation of the robot. Section \ref{sec:exp} summarises the three experiments conducted to verify the efficacy of the proposed system. Section \ref{sec:con} concludes the outcomes of the proposed work suggesting future developments needed to optimize the proposed system. 

\section{Related Work}
\label{sec:bgr}

Research into vision-based navigation systems for crop row following is an extensively explored subject area~\citep{wang2022applications, shalal2013review}. Existing robot navigation technologies for agri-robotics research include GNSS, inertial navigation systems (INS) and light detection and ranging (LiDAR)~\citep{man2020research}. Each of these existing systems presents a cost-benefit trade-off which results in a lack of reliable navigation options for an affordable solution in crop row navigation. The vision-based infield navigation technologies available for arable field navigation explores the usage of RGB and depth images to identify the crop rows~\citep{bonadies2019overview, bai2023vision}.  The existing systems for such infield navigation mainly uses image segmentation, object detection or image matching methods to identify the crop rows for robot navigation~\citep{wang2022applications}. Majority of the existing studies on vision-based infield navigation focuses on the row following aspect of infield navigation~\citep{martini2022position, chaumette2016visual}. Relatively less work has been done towards enabling vision-based crop row switching and headland turning for infield navigation~\citep{wang2022applications}. Switching from one crop row to another often remains unsolved using computer vision~\citep{winterhalter2021localization}. The vision-based crop row switching system demands several perception capabilities including end of row (EOR) detection, re-entry point identification and localisation within the headland~\citep{kanagasingham2020integrating, wang2022applications, li2018image}. 

End of row (EOR) detection is an important step for any crop row switching algorithm since it serves as the starting point for any crop row switching manoeuvre~\citep{de2022vision}. EOR detection was implemented in the vision-based cotton row following algorithm by~\citet{li2018image}. However, their system requires a driver to take over the control during the row-switching stage. This method of manual row-switching behaviour within autonomous row-following systems is a repeating trend in several other infield navigation schemes~\citep{zhang2020cut}. A vision-based EOR detection algorithm was developed based on the percentage of vegetation pixels in the image~\citep{bengochea2016merge}. This method is limited to the headland without any vegetation while it will fail to detect verdant headlands based on a complete vegetation detection-based approach. They also use GNSS-based navigation to execute the row switching behaviour which would require the robot to include a GNSS receiver. The low-cost GNSS receiver used in~\citep{kneip2020crop} robot caused the system to guide the robot towards an undesired crop row due to limited GNSS accuracy. An accurate RTK-GPS system would be required for such GNSS-based systems to perform crop row switching with high fidelity. Some systems rely on GNSS to locate the end of row position along with other sensors such as IMU and compass~\citep{kanagasingham2020integrating, yin2018development}. The EOR detection scheme proposed in~\citep{huang2021row} employs image binarisation using classic computer vision methods to calculate the pixel count in binary masks to determine the EOR. However, the pixel count thresholds may vary depending on the stage of the plant growth and these thresholds must be matched to each growth stage when used for the long term. The pixel counts of a binary image representing the crop would not yield the spatial localisation of EOR within a given image, while it only generates an EOR trigger based on the image. In contrast, the EOR detection method proposed by authors of~\citep{zhang2020cut} uses $Cr$ channel of $YCbCr$ colour space to calculate the position of the EOR within a given image. Such methods pose the advantage of early detection and potential failures due to noisy images in the method presented in~\citep{huang2021row}. However, such colour-based EOR detection methods are highly susceptible to distortions caused by external field variations. The advantages of deep learning-based methods in EOR detection outperform such colour-based methods~\citep{de2023deep, de2022vision}. Lidar and ultrasonic ranging sensors are also used for EOR detection, particularly in the vineyard and orchard navigation scenarios where the absence of adjacent tree rows indicate EOR~\citep{rovira2020augmented, subramanian2008headland}. A 3D point cloud-based row-end detection was introduced in~\citep{kneip2020crop} where the EOR is identified by detection of height drop between the plant and the ground within the point cloud. This approach is mostly limited to crops with noticeable height differences relative to the ground level or crops at later growth stages. The EOR detection algorithm referenced throughout this paper is introduced from our previous work~\citep{de2022vision}, considering the limitations of the existing EOR detection methods. 

Identification of the relative distance between two crop rows is also vital for accurate crop row switching. The distance between two crop rows often referred to as "inter-row distance" is considered as a fixed distance in existing crop row switching methods~\citep{ahmadi2020visual, guevara2020headland}. The relative position between the current robot pose and the next crop row could vary due to imperfections in planting or due to a slight offset during robot navigation. Therefore, active perception of the relative position of the re-entry point to the next crop row to be traversed is a useful attribute. The re-entry point detection algorithm proposed in this work estimates the relative distance to the next crop row based on crop row segmentation mask from Triangle Scan Method (TSM)~\citep{de2023deep} and depth data. Such active perception of inter-row space helps to eliminate any potential row-switching failures caused by varying inter-row space or inaccurate positioning of the robotic platform within the currently traversing crop row.

Despite the considerable interest in vision-based crop row following systems, the presence of reliable vision-based crop row switching algorithms remains limited~\citep{huang2020feedforward}. Most of the existing vision-based navigation systems depend completely or partially on GNSS, INS or Lidar-based solutions for crop row switching rather than a fully vision-based solution~\citep{wang2022applications}. The crop row switching algorithms proposed in~\citep{huang2020feedforward, he2023dynamic, paraforos2018automatic} completely rely on the RTK-GNSS-based sensors to perform the crop row switching manoeuvre. However, GNSS-based systems are not considered a simple and straightforward solution for agricultural robot navigation since they also need multiple redundancies in place for effective operation due to multi-path reflections and signal blockage\citep{rovira2015role}. Some systems also use manual control to perform the headland turn~\citep{kneip2020crop}. Fully vision-based solutions in agricultural robot navigation depend on multiple cameras on the robot for maintaining the localisation during the row switching process~\citep{ahmadi2021towards, liu2023vision}. The methods that use a single camera also demand special requirements such as variable field of view~\citep{xue2011agricultural}. Infield navigation algorithms that completely rely on vision sensors use image processing techniques such as local feature matching~\citep{ahmadi2020visual} and vegetation density thresholding~\citep{xue2012variable}.\citet{ahmadi2020visual} have presented a visual navigation framework for row crop fields. The system only uses onboard cameras for navigation without performing explicit localization. The switching from one row to another is also implemented in the same framework. They have identified that the agricultural environments are lacking distinguishable landmarks to support localisation and mapping, which causes high visual aliasing. They also state that the constantly growing crops make it difficult to maintain a fixed map of the environment. Their robot could: autonomously navigate through row crop fields without maintaining any global reference maps, and monitor crops in the fields with high coverage by accurately following the crop rows, and the system is robust to fields with various row structures and characteristics. The robotic platform was equipped with two cameras (front and back) to perform the crop row switching. The row-switching method employs feature matching to traverse towards the next crop row. Their work indicates that the vision-based navigation in crop rows could be performed without the need for a global map. The importance of the field of view (FOV) of the cameras used in vision-based navigation systems was indicated by~\citet{xue2012variable} their work on crop row navigation. The researchers used a single monocular camera with a variable FOV setup for better accuracy in navigation at the end of crop rows. Thus the researchers identified the requirement of a wider field of view towards the success of vision-based crop row navigation. The lack of research in fully vision-based crop row switching approaches with vision-based EOR detection indicates a clear gap in the literature within the vision-based infield navigation domain. The four most common headland turning patterns are semi-circular turn, U-turn, light bulb turn (a.k.a $\Omega$ turn) and switch-back turn~\citep{he2023dynamic}. The semi-circular turn describes a half circle with a constant radius while the U-turn describes two-quarter circle turns ($90^{\circ}$) centred with a linear traversal stage. These methods are typically used in smaller robots with tighter turning radii relative to the inter-row distance~\citep{huang2021row, ahmadi2021towards, guevara2020headland}. The $\Omega$ turn and the switch-back turn are mostly used in robots with constrained manoeuvrability such as Ackerman steering robots or tractors~\citep{he2023dynamic, wang2018adaptive, evans2020row}. Considering the skid-steering robot configuration of the robotic platform used in this work, the U-turn pattern was chosen to execute the row-switching manoeuvre.

\section{Vision-based Navigation in Arable Fields}
\label{sec:vbaf}
The vision-based navigation pipeline we propose only uses RGB images for in-row navigation and it uses RGB-D images for row switching. This system does not rely on any assumptions related to the environment it's deployed, such as type of crop, inter-row space, plant size, sunny or overcast conditions, soil texture and other field variations identified in~\citep{de2023deep}. An outline of the hardware platform used and the generalization ability of the proposed system to other robotics platforms is presented below. The overall navigation paradigm is also presented in this section. 

\subsection{Hardware Requirements and Interoperability}
\begin{figure}
\centering
\captionsetup{justification=raggedright,singlelinecheck=false}
\includegraphics[scale=0.45]{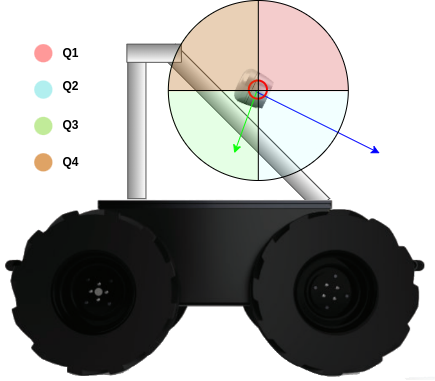}
\caption{Camera positioning for Triangle Scan Method. The principal axis of the camera(blue) must always reside within Q2.}
\label{fig:qd}
\end{figure}

The proposed system completely relies on RGBD data from a stereo camera. The camera was mounted onto a mobile robotic platform while its principle axis lies in the $Q2$ quadrant of the vertical plane along the centre of the robot as illustrated in Figure \ref{fig:qd}. The crop row detection algorithm remains unaffected by the camera oscillations within the $Q2$ quadrant. This navigation system demands a robotic platform which provides wheel odometry. Platforms which provide IMU-corrected wheel odometry provide better performance in crop row switching steps within uneven headland areas.

\begin{figure}
\centering
\captionsetup{justification=centering}
\includegraphics[scale=0.6]{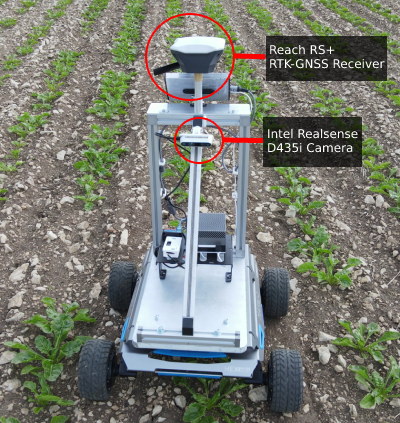}
\caption{Hexman Mark-1 robot in the Sugar Beet Field.}
\label{fig:m1}
\end{figure}

The proposed system could be deployed on any mobile robotic platform employing an RGBD camera with correct placement as explained above. The crop row detection algorithm could be tuned for optimal operation by adjusting the pitch angle of the camera within $Q2$ followed by the suggested camera placement. A calibration program will lay out virtual guidelines on the camera image frame for this adjustment. A detailed calibration guide on camera placement is presented in our previous work~\citep{de2022vision}. This system was deployed on two robotic platforms: Clearpath Husky and Hexman Mark-1. However, Hexman Mark-1 robot was used for the experiments outlined in Section \ref{sec:exp} due to its longer battery life.  Figure \ref{fig:m1} shows the robotic setup with a front-mounted Intel RealSense D435i camera and a Reach RS+ RTK GNSS receiver. An NVIDIA Jetson AGX Orin developer kit was used as the onboard computer for the robot. Mark-1 is a skid steer robot with a ground clearance of 128 mm. The dimensions of the robot are $526\times 507\times 244$ mm.

\subsection{Infield Navigation Scheme}
\label{sec:ins}
\begin{figure}
\centering
\captionsetup{justification=centering}
\includegraphics[scale=0.19]{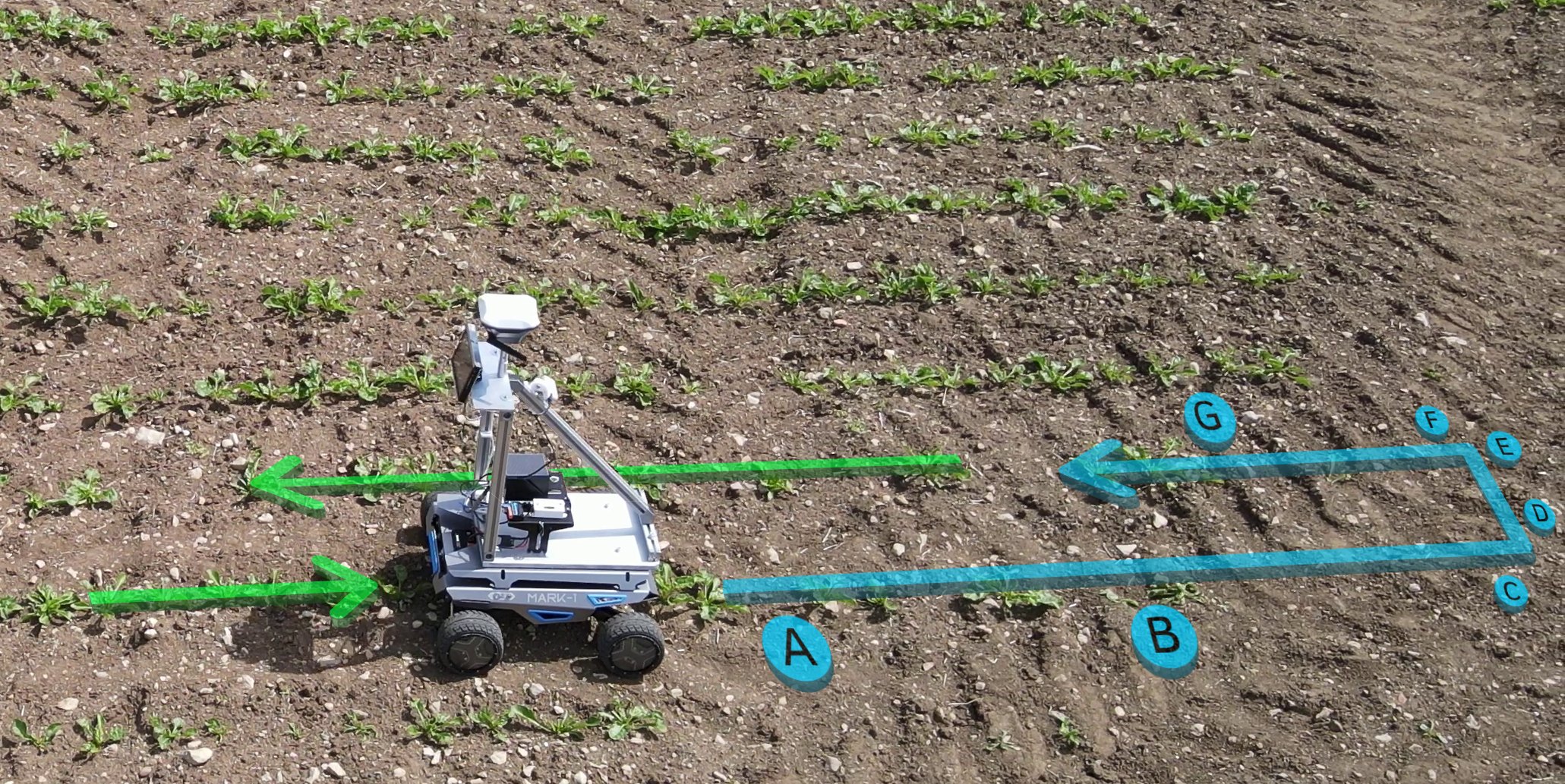}
\caption{Vision-based infield navigation scheme. Green: In-row navigation behaviour, Blue: Row switching behaviour.}
\label{fig:utr}
\end{figure}

The infield navigation scheme comprises two main components in its navigation behaviour. In-row navigation component~\citep{de2022vision} is a reactive navigation strategy which uses RGB images from a single front-mounted camera. The captured images are used to detect the angle and linear offset of the central crop row relative to the robot. The angular velocity of the robot is controlled based on these central crop row parameters. The row switching component introduced in Section \ref{sec:rsw} detects the end of the crop row (EOR) and re-entry point to the next row based on RGB images. An aligned depth map of the RGB image was used to identify the distance offset to the next row. This distance offset and EOR position is used in a path planner which executes a U-turn manoeuvre in the form of a 7-step state machine. Figure \ref{fig:utr} illustrates the process of the entire infield navigation scheme. The in-row navigation regions are marked with green arrows and the row switching manoeuvre is illustrated in blue. 

This infield navigation scheme combines four technical components: a vision-based crop row detection method, a crop row following algorithm based on the detected crop rows, a row switching manoeuvre which perpetuates the row following behaviour culminating an infield navigation scheme and a initial turning direction detection algorithm to initiate the navigation direction. Each of these components are elaborated in the following sections. 

\subsection{Crop Row Detection}
\begin{figure}
\centering
\captionsetup{justification=raggedright,singlelinecheck=false}
\includegraphics[scale=0.4]{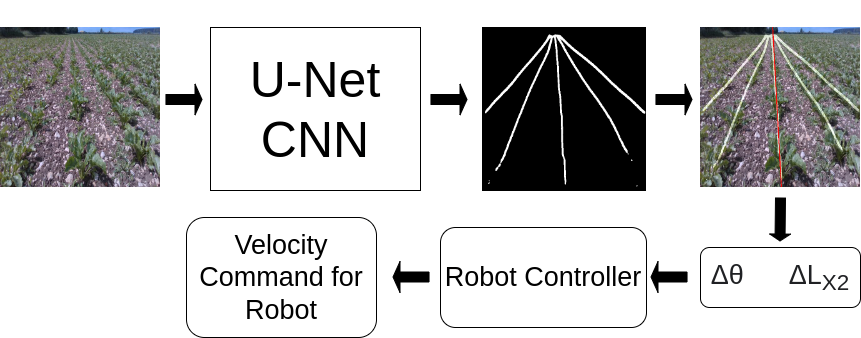}
\caption{The proposed crop row following architecture with U-Net CNN for crop row mask detection. The crop mask generated by U-Net CNN is used by a triangle scan method to predict a central crop row ($\Delta \theta$: Crop row angle error corresponding to vertical axis, $\Delta L_{x2}$: Positional error of the central crop row relative to image midpoint).~\citep{de2023deep}}
\label{fig:ovl}
\end{figure}

The crop row detection sub-component of this algorithm uses the U-Net~\citep{ronneberger2015u} based semantic segmentation to generate a crop row mask of the RGB image received from the robot camera. The crop row mask predicted by the U-Net CNN is characterised by a skeleton representation of the crop row rather than attempting to segment the entire region of the crop row within the image as shown in Figure \ref{fig:ovl}. This skeleton representation enables the model to generalize to most of the crops without explicit crop-specific training of the CNN model. 

\begin{figure}
\centering
\captionsetup{justification=raggedright,singlelinecheck=false}
\includegraphics[scale=0.4]{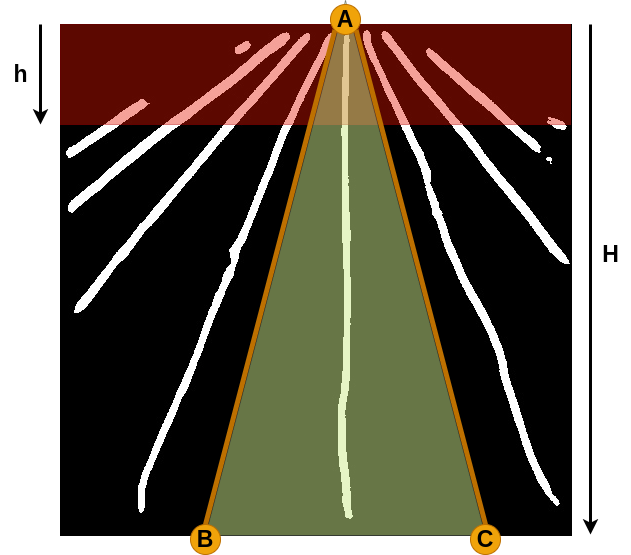}
\caption{Regions of interest for scanning the top and bottom points of central crop row. Vanishing point scan ROI: RED, Bottom point scan ROI: Green, H: Height of the image, h: Vanishing point scan ROI height.~\citep{de2023deep}}
\label{fig:roi}
\end{figure}

The triangle scan algorithm (TSM)~\citep{de2023deep} is a post-processing algorithm for the predicted crop row mask which extracts the central crop row line parameters(position and orientation). The TSM algorithm identifies the horizontal position of the vanishing point for all crop rows within the crop row mask by analyzing a narrow strip of pixels at the top of the image. An isosceles triangle ROI is defined with this point being the top position of the triangle and the other two points residing on the bottom edge of the image. The bottom position of the central crop row is scanned within this triangle. These ROIs are illustrated in Figure \ref{fig:roi}. Angular position ($\Delta\theta$) and displacement ($\Delta L_{x2}$) error extracted by the TSM was used for crop row following. 

\subsection{Crop Row Following}
\label{sec:crf}
The output of the TSM algorithm is used as the input parameters for a robotic controller which steers the robot while it straddles the crop row as shown in Figure \ref{fig:utr}. A proportional, integral and derivative (PID) controller was used with platform-specific tuning for the robotics platform used in our experiments. This controller stage is considered as a modular entity which can be replaced with any other desired controller. The bipartite crop row error characterised by the TSM is integrated to a single error value $E$ by calculating a weighted sum of the error terms with weights $w_{1}$ and $w_{2}$ as described in Equation \ref{eq:er}. This composite error $E$ could be fed into a desired robotic controller to generate an angular velocity signal $\omega$ which would directly control the robot heading relative to the crop row. This control paradigm is best identified as an image-based visual servoing (IBVS) technique~\citep{chaumette2016visual} where the robot is controlled to minimize an error directly extracted from an image. 

\begin{equation} \label{eq:er}
  {E} =  w_{1}\Delta\theta + w_{2} \Delta L_{x2} 
\end{equation}

The crop row following algorithm was executed in a real sugar beet field with varying initial angular errors ranging from -13.53$^\circ$ to 33.14$^\circ$. The tuned PID controller could bring down the error to 80\% to 90\% of the initial error within 4 meters of traversal into the crop row. The long-distance navigation experiment outlined in Section \ref{sec:ex1} will investigate the accuracy of a row following over long distances while navigating crop rows in a real sugar beet field. 

\subsection{Crop Row Switching}
\label{sec:rsw}
\begin{figure}
\centering
\captionsetup{justification=centering}
\includegraphics[scale=0.37]{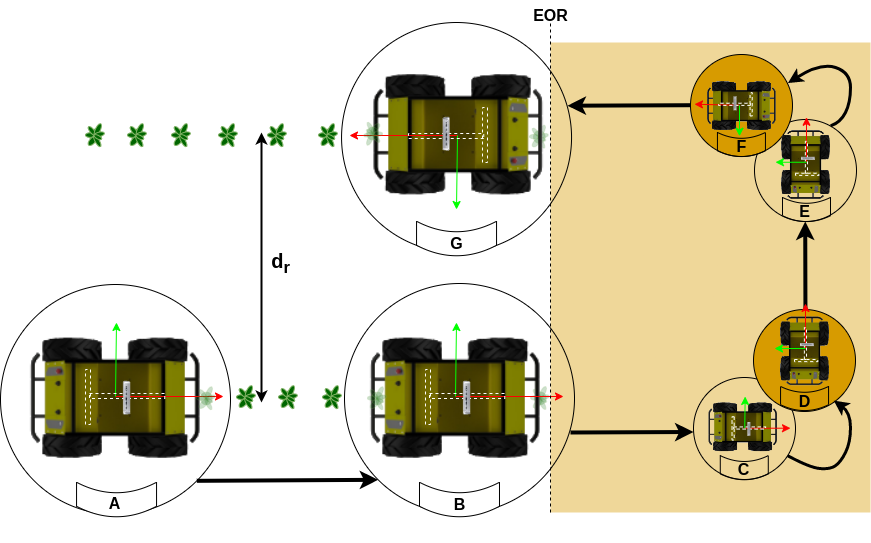}
\caption{Crop row switching state machine. $d_{r}$: Distance between current and next crop row, A: Initial Detection of the EOR, B: Robot is at the EOR, C: Robot traversed a distance equal to its length into the headland, D: Robot turn 90$^\circ$ towards the next row direction from state C, E: Robot traverse $d_{r}$ distance forward from state D, F: Robot turn 90$^\circ$ towards the next row direction from state E, G: Robot re-enter the next crop row. }
\label{fig:fsm}
\end{figure}

Figure \ref{fig:fsm} portrays a state machine representing the process of crop row switching, spanning across a total of seven states. Table \ref{tab:stt} describes each state of the process given in Figure \ref{fig:fsm} encountered during the row switching manoeuvre. The blue colour overlay in Figure \ref{fig:utr} illustrates the crop row switching manoeuvre executed by the robot upon detection of the EOR (state $A$). The robot uses the EOR detector described in~\citep{de2022vision} to detect the EOR while traversing the crop row. The robot is switched back to row-following mode at the end of the row-switching manoeuvre (state $G$). The switching manoeuvre is composed of three steps: row exit, U-turn and re-entry. The transitions $A\rightarrow B\rightarrow C$ belong to the step of row exit. The U-turn step contains the transitions $C\rightarrow D\rightarrow E\rightarrow F$ while the transition $F\rightarrow G$ is considered as a re-entry step. The methods and techniques used in each of these three steps are explained in sections \ref{sec:exit}, \ref{sec:ut} and \ref{sec:re} respectively.

The experiment on row switching manoeuvre outlined in Section \ref{sec:ex2} reported a lower success rate due to the slipping of the robot within the headland area during the 90$^\circ$ turns ($C\rightarrow D$ and $E\rightarrow F$) of the row switching manoeuvre. The row-switching algorithm is improved by adding translational odometry tracking to the state transitions $C\rightarrow D$ and $E\rightarrow F$ to accurately traverse the inter-row distance $D\rightarrow E$. The impact of this improvement is reported in the experimental results reported in Section \ref{sec:ex3}.

\begin{table}
\centering
\caption{States encountered during crop row switching manoeuvre.}
\begin{center}
\begin{tabular}{|p{0.07\linewidth} | p{0.8\linewidth} |}
\hline
\textbf{State} & \textbf{Description} \\
\hline
$A$ & Position of the robot during initial EOR detection.  \\ 
\hline
$B$ & Robot reaches the EOR position.  \\ 
\hline
$C$ & Robot enters the headland area completely passing EOR position.   \\ 
\hline
$D$ & Robot turns 90$^\circ$ towards the direction of the next crop row from state $C$.  \\ 
\hline
$E$ & Robot aligns itself with the next crop row traversing forward from state $D$.  \\ 
\hline
$F$ & Robot turns 90$^\circ$ facing itself towards the starting point of the next crop row while residing within the headland buffer.  \\ 
\hline
$G$ & Robot enters the next crop row.  \\ 
\hline
\end{tabular}
\label{tab:stt}
\end{center}
\end{table}

\subsubsection{Row Exit Step}
\label{sec:exit}
Row exit is the process in which the robot drives itself completely out of the currently traversing crop row after detecting the EOR. The EOR is initially detected at state $A$ where the robot estimates the relative 3D coordinate of the starting point of the next crop row the robot would enter. The Y value of this relative 3D coordinate would represent the inter-row distance $d_{r}$ between the current and the next crop row. The $d_{r}$ value is useful when the robot is traversing during the $D\rightarrow E$ transition. The robot transits through the states $A\rightarrow B\rightarrow C$ after the re-entry point detection using a combination of visual feature matching and odometry as explained in the following subsections. 

\noindent\textbf{Re-entry Point Detection:}

\begin{figure}
\centering
\captionsetup{justification=centering}
\includegraphics[scale=0.45]{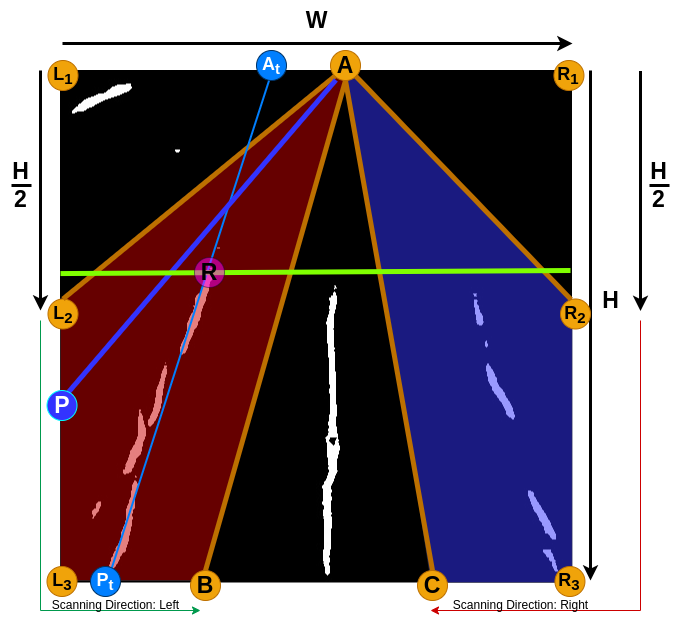}
\caption{Regions of interests (ROI) for re-entry point scanning. Red: Left side ROI, Blue: Right side ROI, Green: Detected EOR line.}
\label{fig:roi2}
\end{figure}

The re-entry point detection module is extended based on the TSM crop row detection pipeline from our previous work~\citep{de2023deep} on crop row detection. Similar to TSM, a deep learning-based skeleton segmentation of crop rows was used as the input for the re-entry point detector. As illustrated in Figure \ref{fig:roi2}, the ROI $AL_{2}L_{3}B$ was defined if the next intended turn is to the left ($AR_{2}R_{3}C$ for right). Points $A,B$ and $C$ were determined using the "Anchor scans" step of TSM~\citep{de2023deep}. The horizontal green line in Figure \ref{fig:roi2} was obtained using the EOR detector~\citep{de2022vision}. Equation \ref{eq:1} yields a point $P_{t}$ by scanning for the pixel sum along $AP$ line where point $P$ is an arbitrary point on $L_{2}L_{3}B$ path. Similarly,  equation \ref{eq:2} yields a point $A_{t}$ by scanning the pixel sum along $\overline{A}P_{t}$ line where point $\overline{A}$ is an arbitrary point on $AL_{1}$ line. The intersection between the EOR line and $A_{t}P_{t}$ is identified as the re-entry point $R$ for the next crop row. The depth information from the corresponding depth image was used to determine the 3D coordinate corresponding to the point $R$ which was then used to determine $d_{r}$.

\begin{equation} \label{eq:1}
  P_{t} = \argmax \Biggl[ \sum_{I_{xy}=A}^{P} I(x,y) \Biggr]_{P=L_{1}\rightarrow L_{2}}^{P=L_{2}\rightarrow B}
\end{equation}

\begin{equation} \label{eq:2}
  A_{t} = \argmax \Biggl[ \sum_{I_{xy}=P_{t}}^{\overline{A}} I(x,y) \Biggr]_{\overline{A}=A}^{\overline{A}=L_{1}}
\end{equation}

\noindent\textbf{A to B Transition:}

The TSM-based crop row navigation framework was proven to be able to maintain the average heading of the robot relative to the crop row under 1$^\circ$~\citep{de2022vision} . Therefore, the relative heading angle between the robot and the crop row was assumed to be under 1$^\circ$ at state $A$. The detected EOR line in Figure \ref{fig:roi2} demarcates the headland area and the crop row region within the RGB image obtained from the front-mounted camera. The RGB image was cropped below the EOR line and saved as a reference image $I_{R}$ at state $A$. The robot was then moved towards the EOR with a constant forward linear velocity while calculating the local feature (Scale Invariant Feature Transform~\citep{lowe2004distinctive} was used) similarity score in each new image captured by the robot camera with $I_{R}$. The robot was stopped and assumed to reach state $B$ when the feature similarity score dropped below an experimentally determined threshold. The threshold value was determined by observing the feature similarity score while driving the robot to the actual EOR position using teleoperation.

\noindent\textbf{B to C Transition:}

The frontmost edge of the robot is coincident with EOR at state $B$. The minimum distance the robot must move in order to completely exit the crop row is the length of the robot ($L_{R}$) itself. The wheel odometry of the robot was used as feedback to move the robot forward towards the headland. The length of the Hexman Mark 1 robotic platform used in during the field trials was 526 mm. The wheel odometry of the robot was assumed to be accurate enough to navigate the robot forward into the headland at such small distances. 

\subsubsection{U-Turn Step}
\label{sec:ut}
The U-turn step involves the robot taking two 90$^\circ$ turns with a linear navigation stage ($D\rightarrow E$) in between. "Headland buffer" region is the space in the headland that is directly in front of a given crop row bound by the EOR, edge of the field and the two centre lines of the inter-row space between adjacent crop rows. The goal of the U-turn step is to bring the robot into the headland buffer region of the next crop row while facing towards the next crop row to be traversed. It was experimentally verified that the TSM can resume its normal crop row navigation framework from this point(state $F$) onward to traverse in to the crop row. The 90$^\circ$ turns and $D\rightarrow E$ transition is executed with wheel odometry feedback of the robot. The practical behaviour of the robot during rotation stages ($C\rightarrow D$ and $E\rightarrow F$) ensured that the robot would reach the headland buffer when the $D\rightarrow E$ transition distance was set to $d_{r}$.

\subsubsection{Re-Entry Step}
\label{sec:re}
The robot is within the headland buffer region at state $F$ and state $G$ is reached when the robot enters the next crop row to be traversed. The re-entry step is the transition from state $F$ to state $G$ where the robot moves from the headland buffer into the crop row in front of it. This transition was realized by launching the TSM-based crop row following framework at state $F$. The TSM was able to detect the crop row in front of the robot and navigate the robot into the crop row. This behaviour of TSM-based re-entry navigation is verified by the experiment outlined in Section \ref{sec:rev}.

\subsection{Initial Turning Direction Detection}
\label{sec:vbifo}

\begin{figure}
\centering
\captionsetup{justification=raggedright,singlelinecheck=false}
\includegraphics[scale=0.4]{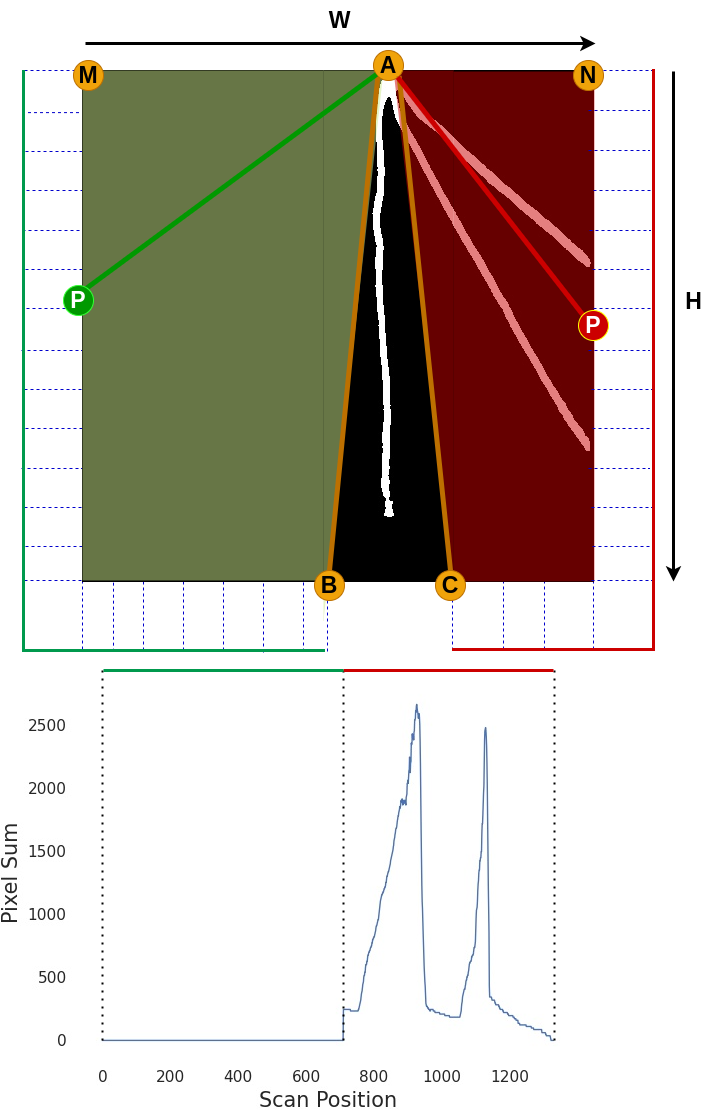}
\caption{Graphical representation of initial turning direction detection calculation. Green: ROI to the left of the central line, Red: ROI to the right of the central line, H: Height of the image, W: Width of the image.}
\label{fig:froi}
\end{figure}

Row following and row switching behaviours explained in Sections \ref{sec:crf} and \ref{sec:rsw} must be alternated to realise a field scale navigation scheme. Assuming the robot is starting at an edge of the field (left or right), the initial turning direction of the robot at the first-row switching instance must be determined to propagate the field scale navigation scheme in the desired direction. The asymmetric crop row distribution in the predicted crop row mask at the crop rows next to an edge of the field could be exploited to identify the initial turning direction. The predicted crop row mask shown in Figure \ref{fig:froi} is from a crop row near the left edge of the sugar beet field, in which there are multiple crop rows detected to the right of the central crop row and none to the left. This asymmetric crop row distribution could be formalised by calculating the ratio of maximum sweeping pixel sums to the left and right of the central crop row predicted by the triangle scan algorithm as shown in Figure \ref{fig:froi}. The sweeping pixel sum of the crop row mask along the AP line where P is a variable point on the image border sections MB (\rotatebox[origin=c]{180}{$\Lsh$}) and CN (\rotatebox[origin=c]{270}{$\Lsh$}) are represented by the green and red line segments respectively in Figure \ref{fig:froi}. This pixel sum variation is plotted in the graph at the bottom of Figure \ref{fig:froi} while P changes from M to B and C to N. The ratio of peak values of the sum of the pixels in the MB section (green) and CN section (red), denoted by $O_{F}$ could be used as an indicator to identify asymmetric crop row distributions as outlined in Equation \ref{eq:ori}. An empirical threshold value was determined to identify the initial turning direction as indicated in Equation \ref{eq:dir}. The robot is considered to be in a crop row in the middle of the field if none of the criteria outlined in Equation \ref{eq:dir} is met. The heuristics outlined by Equations \ref{eq:ori} and \ref{eq:dir} could be used to determine both the initial turning direction and as a signal to identify the end of the field when the robot is traversing the last crop row in a field. 

\begin{equation} \label{eq:ori}
  O_{F} = \frac{\argmax \Biggl[ \sum_{I_{xy}=A}^{P} I(x,y) \Biggr]_{P=M}^{P=B}}{\argmax \Biggl[ \sum_{I_{xy}=A}^{P} I(x,y) \Biggr]_{P=C}^{P=N}}
\end{equation}

\begin{equation} \label{eq:dir}
  Initial \ Turning \ Direction =
  \begin{cases}
    left, & \text{if } O_{F} > 6, \\
    right, & \text{if } \frac{1}{O_{F}} > 6.
  \end{cases}
\end{equation}

\section{Experimental Study}
\label{sec:exp}
Three experiments were carried out to evaluate the robustness of the proposed vision-based navigation system. The first experiment will evaluate the row-following capability of the system over long distances under varying field conditions. The second experiment examines the efficacy of crop row switching manoeuvre to navigate the robot towards the next crop row to be traversed. The third experiment will analyse the effect of the crop row switching mechanism towards the overall field scale navigation system. These three experiments will demonstrate the performance of the key components of this system: row following, row switching and the ability to scale up the row following capability to achieve field scale navigation with row switching. 

\subsection{Experiment 1: Long Distance Navigation}
\label{sec:ex1}
This experiment was devised to examine the performance of the proposed crop row following algorithm over long distances. A circuit of 10 crop rows was selected to test the robot's navigation using the vision based navigation system. The robot navigated autonomously through this circuit twice, during the early growth stage and the later growth stage. 

\subsubsection{Ground truth crop row positions}
\begin{figure}
\centering
\captionsetup{justification=raggedright,singlelinecheck=false}
\includegraphics[scale=0.13]{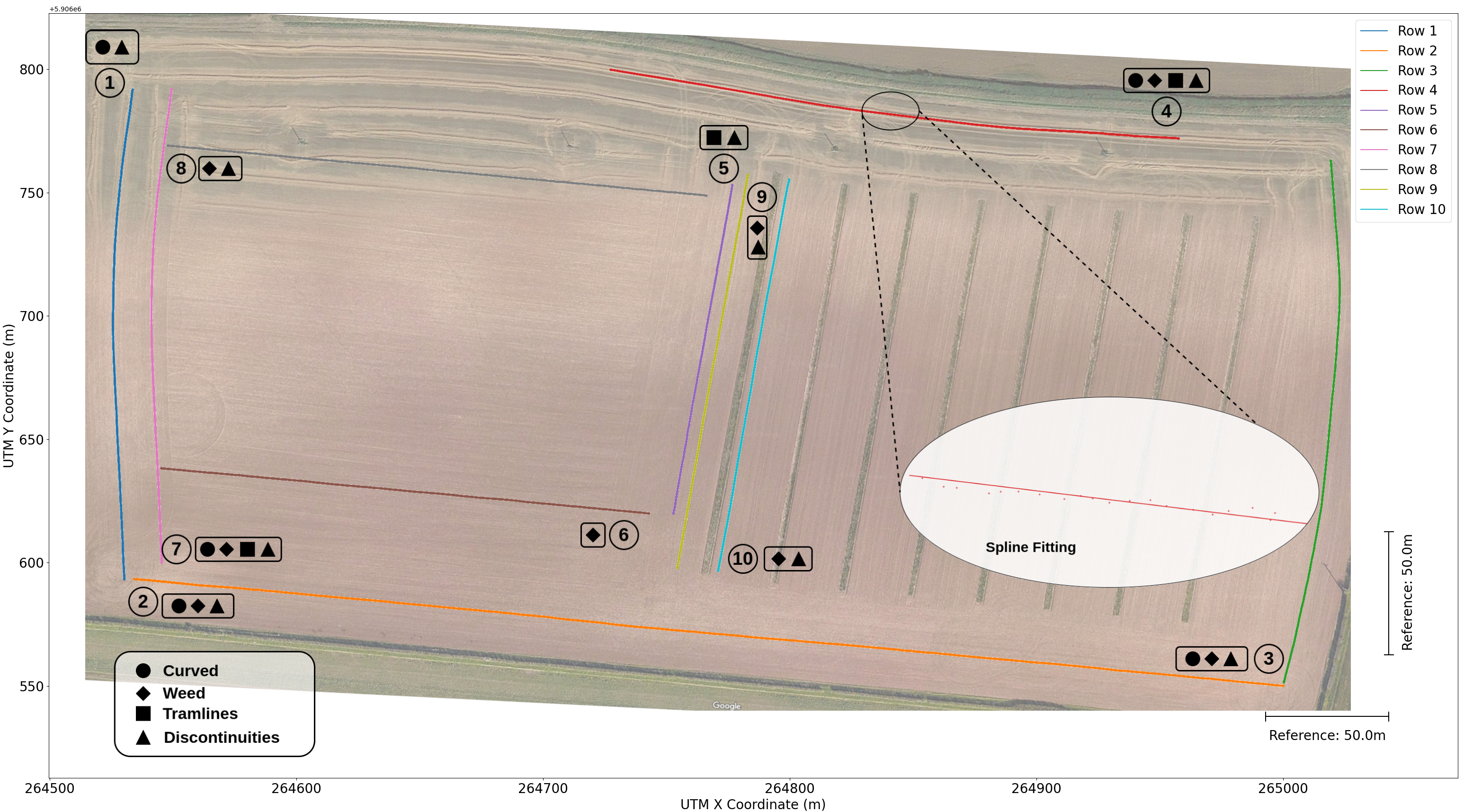}
\caption{Selected crop rows for long-distance navigation experiment with field variations: curved crop rows [$\bullet$], weed presence [$\blacklozenge$], tramlines [$\blacktriangle$] in image frame and discontinuities due to missing plants[$\blacksquare$]. $[1\downarrow] , [\overrightarrow{2}] , [3\uparrow] , [\overleftarrow{4}], [5\downarrow] , [\overleftarrow{6}] , [7\uparrow] , [\overrightarrow{8}] , [9\downarrow] , [10\uparrow]$}
\label{fig:bs}
\end{figure}

Ten crop rows marked in Figure \ref{fig:bs} were selected to evaluate the long-distance navigation capability of the vision based navigation system in a sugar beet field. These ten crop rows had one or more physical field variations: curved crop rows ($\bullet$), weed presence ($\blacklozenge$), tramlines ($\blacktriangle$) in the image frame and discontinuities due to missing plants($\blacksquare$). This field also had an ongoing agroforestry~\citep{ramachandran2009agroforestry} setup with 8 rows of trees planted on the right side of the field. The 10\textsuperscript{th} crop row is positioned right next to one of these agroforestry tree rows. This is an additional abnormality in the field compared to the field environments on which the crop row detection AI model was trained on~\citep{de2023deep}. The row number in Figure \ref{fig:bs} is placed at the starting traversal point of each crop row and the traversal directions encompass all four boundaries of the field. The 10 crop rows were distributed throughout the entire field with a cumulative circuit distance totalling 2.25 km. 

The Hexman Mark-1 robot equipped with an EMLID Reach RS+ RTK GPS was used to record the GNSS coordinated of the 10 crop rows. The horizontal kinematic precision of the EMLID Reach RS+ RTK GPS was 7 mm + 1 ppm (1 mm error per 1km baseline increment).  The robot was driven at very slow speeds (<0.3 ms\textsuperscript{-1}) during the early growth stage of the crop by an expert human driver using a remote controller while the ground truth crop row positions were being recorded.  A third-order polynomial spline was fitted on each of the collected crop row ground truth GNSS coordinate trajectories to create a continuous representation of the crop row ground truth position as denoted in Equation \ref{eq:spl}. The number of GNSS coordinates in each trajectory is n and the spline function  $S(x)$ was created using "UnivariateSpline" method in scipy python library. This cubic polynomial was subjected to a smoothing condition stated in Equation \ref{eq:sm} with a smoothing parameter s(=2). The weight term $w_i$ was set to $1$ during the smoothing condition calculation. The smoothing condition avoids overfitting of the spline effectively rejecting noise in the collected GNSS coordinates. Figure \ref{fig:bs} shows a zoomed in view of the spline fitted among the GNSS coordinates in a selected portion of the 4\textsuperscript{th} crop row.

\begin{equation} \label{eq:spl}
  S(x) = \sum_{i=1}^{n} a_i(x - x_i)^3 + b_i(x - x_i)^2 + c_i(x - x_i) + d_i
\end{equation}

\begin{equation} \label{eq:sm}
  \sum_{i=1}^{n} w_i \cdot (y_i - \text{S}(x_i))^2 \leq s
\end{equation}

\subsubsection{Navigation Evaluation}
\label{sec:ne}

\begin{figure}
\centering
\captionsetup{justification=raggedright,singlelinecheck=false}
\includegraphics[scale=0.25]{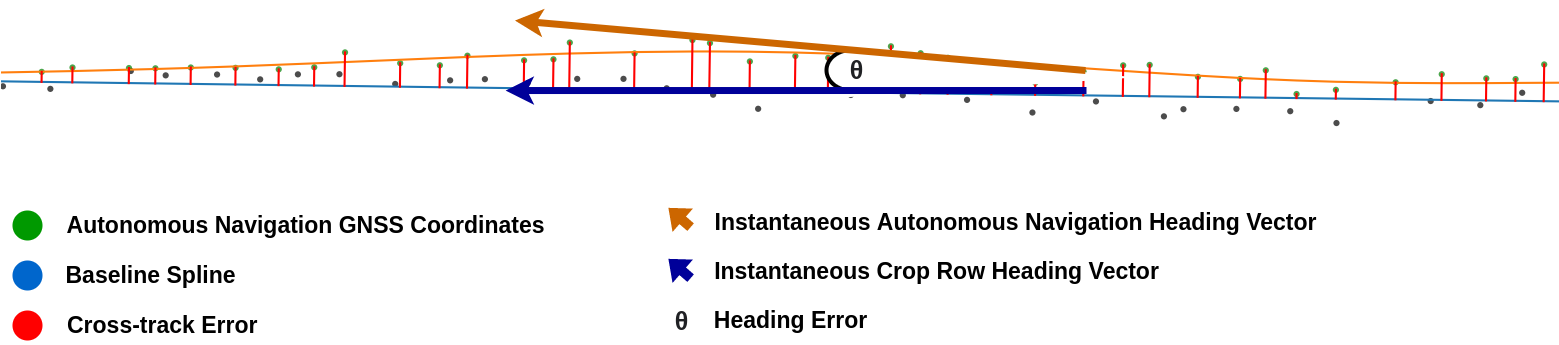}
\caption{Visualisation of cross-track and heading error calculations.}
\label{fig:ers}
\end{figure}

The robot navigation was tested during both early growth stage ($\leq$ 4 leaves per plant) and the late growth stage ($\geq$ 6 leaves per plant). The robot completed the entire 10 crop row circuit during both stages resulting in a total autonomous navigation distance of 4.5 km. The GNSS coordinates of the robot were recorded during each traversal of a crop row and all the points including the ground truth coordinates were converted to Universal Transverse Mercator (UTM) for analysis. The collective root mean square error for the precision of recorded GNSS trajectories amounted to 1.8 cm. The perpendicular distance from each of the recorded points during autonomous navigation to the respective ground truth spline $S_{i}(x)$ of i\textsuperscript{th} crop row was calculated. This perpendicular distance is considered as the offset of the robot from the desired crop row path during autonomous crop row following (cross-track error). The average heading error during navigation was also calculated based on the difference of the instantaneous heading angles of the ground truth spline $S_{i}(x)$ of i\textsuperscript{th} crop row and the autonomous navigation trajectory. The instantaneous heading angles on the autonomous navigation trajectory was calculated by obtaining the angle of the tangent line at each of the GNSS points with respect to a third-order polynomial spline (s=0.5) fitted on the autonomous navigation trajectory based on Equations \ref{eq:spl} and \ref{eq:sm}. The spline fitted on the autonomous navigation trajectory is indicated in orange colour in Figure \ref{fig:ers} where $\theta$ is the instantaneous heading error. 

Table \ref{tab:res} summarises the cross-track error and heading error during the long-distance navigation experiment. Each crop row had a different length and other physical variations as indicated in Figure \ref{fig:bs}. The overall median cross-track was recorded at 3.32 cm while the average was recorded at 3.99 cm. Both median and average heading errors were recorded at 1.24$^\circ$ with the maximum recorded heading error being 3.29$^\circ$ from row 7 during the late growth stage. Heading and cross-track errors from late growth stage navigation recorded to be slightly higher than that of early growth stage navigation. The crop row detections during the early growth stage tend to be closer to the recorded ground truth positions due to the smaller size of the plants. The crop canopy grows larger during the late growth stage and the detected crop row could have a slight offset from the actual emergence point of the plant stem (lateral meristem region). However, these errors are negligible considering the slight difference between early and late growth stage error values. The horizontal kinematic accuracy of the GNSS sensor was 7mm + 1ppm according to the specifications of EMLID Reach RS+ RTK-GNSS module~\citep{kumar2020design}. The 33 mm accuracy recorded on our system is similar to the expected performance of the Reach RS+ RTK GNSS sensor with a base station located 26km away. 

\begin{table}
    \centering
    \caption{Median average cross-track and heading errors during traversal of each crop row}
    \begin{tabular}{|c|c|c|c|c|c|}
        \hline
        \multirow{2}{*}{\textbf{Row Number}} & \multirow{2}{*}{\textbf{Row Length (m)}} & \multicolumn{2}{|c|}{\textbf{Cross-track Error (cm)}} & \multicolumn{2}{|c|}{\textbf{Heading Error ($^\circ$)}} \\
        
         &  & \textbf{Early Stage} & \textbf{Late Stage} & \textbf{Early Stage} & \textbf{Late Stage} \\
        \hline
        1 & 207.52 & 4.73 & 4.30 & 0.42 & 1.51 \\
        2 & 482.11 & 2.85 & 2.48 & 1.43 & 3.21 \\
        3 & 217.05 & 3.13 & 3.06 & 0.49 & 1.41 \\
        4 & 238.14 & 3.85 & 5.02 & 0.26 & 1.24 \\
        5 & 140.50 & 3.12 & 3.06 & 1.72 & 1.05 \\
        6 & 204.08 & 2.56 & 2.47 & 2.57 & 0.74 \\
        7 & 202.61 & 4.19 & 3.44 & 0.81 & 3.29 \\
        8 & 225.51 & 2.75 & 4.08 & 1.13 & 0.03 \\
        9 & 167.42 & 2.62 & 3.15 & 1.98 & 0.98 \\
        10 & 163.70 & 2.60 & 2.86 & 0.26 & 0.28 \\
        \textbf{Average} & \textbf{224.86} & \textbf{3.24} & \textbf{3.39} & \textbf{1.11} & \textbf{1.37} \\
        \hline
    \end{tabular}
    \label{tab:res}
\end{table}

\subsubsection{Field Variation Analysis}
\label{sec:fva}

The results of the autonomous navigation trials are re-evaluated based on the individual field variations available in the traversed crop rows in Table \ref{tab:fv}. The slightly higher error values were observed in the late growth stage crop row navigation compared to the early growth stage navigation trials in this analysis as well. There was a clear difference in both heading and cross-track errors for crop rows that were curved and positioned near tramlines. The system performance has not been significantly influenced by the presence of weeds and missing plants in crop rows. The initial stress testing experiments of the proposed visual servoing controller concluded that the robot could be brought back into the crop row as long as the robot stays within 20$^\circ$ heading deviation from the crop row~\citep{de2022vision}. The results of this long-distance navigation experiment show that the robot's heading deviation stays under 4$^\circ$ during autonomously navigating a 4.5 km distance in an arable field. To this end, this experiment reinforces the promise of the proposed vision-based crop row following pipeline to be well within the tested performance margins of the employed controller. The previous experiments conducted on vision-based crop row detection~\citep{de2023deep} concluded that the proposed perception algorithm can accurately predict the crop row positions under varying field conditions. The long-distance navigation experiment presented in this paper further confirms the robustness of our crop row detection algorithm as a reliable perception method for crop row following under varying field conditions through different growth stages of the crop. 

\begin{table}
    \centering
    \caption{Median average cross-track and heading errors for each physical crop row variation}
    \begin{tabular}{|c|c|c|c|c|c|c|}
        \hline
        \multirow{2}{*}{\textbf{Row Variation}} & \multicolumn{2}{|c|}{\textbf{Cross-track Error (cm)}} & \multicolumn{2}{|c|}{\textbf{Heading Error ($^\circ$)}} & \multicolumn{2}{|c|}{\textbf{Overall Error}} \\
        
         & \textbf{Early Stage} & \textbf{Late Stage} & \textbf{Early Stage} & \textbf{Late Stage} & \textbf{Cross-track (cm)} & \textbf{Heading ($^\circ$)} \\
        \hline
        Curved & 3.75 & 3.66 & 0.68 & 2.13 & 3.71 & 1.41\\
        Weed & 3.07 & 3.32 & 1.12 & 1.40 & 3.20 & 1.26\\
        Tramlines & 3.72 & 3.84 & 0.93 & 1.86 & 3.78 & 1.39\\
        Discontinuities & 3.32 & 3.50 & 0.95 & 1.44 & 3.41 & 1.20\\
        \hline
    \end{tabular}
    \label{tab:fv}
\end{table}

\subsubsection{Discussion}
\begin{figure}
\centering
\captionsetup{justification=raggedright,singlelinecheck=false}
\includegraphics[scale=0.29]{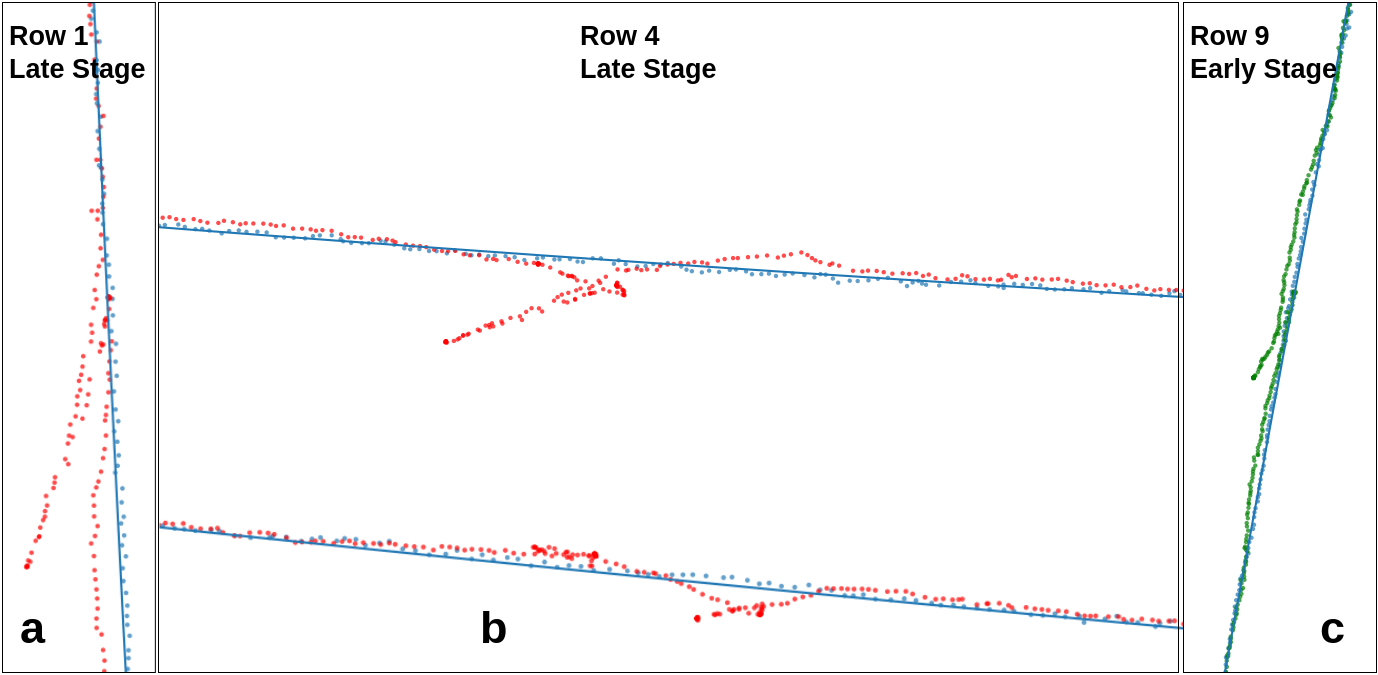}
\caption{Instances where the robot left the traversing crop row failing to recover itself towards the desired path. a: Due to a pothole in the inter-row space, b, c: Due to the large discontinuities in crop row.}
\label{fig:ro}
\end{figure}

There were four instances where the robot completely deviated from the traversing crop row, moving towards the adjacent crop row during the 4.5km autonomous navigation trials. The human driver overtook the control of the robot in these instances and brought the robot back into the traversing crop row in all these instances as seen in Figure \ref{fig:ro}. Each of the occurrences exposed certain limitations of the hardware and software aspects of the proposed approach. The instance \textbf{a} in Figure \ref{fig:ro} caused due to one of the wheels of the robot being trapped in a pothole while traversing the crop row. As seen in Figure \ref{fig:m1}, the presence of large rock fragments is a salient feature in the sugar beet field where these experiments were conducted. While potholes are a rare occurrence in arable fields, this scenario was caused by the displacement of a large rock in the inter-row space. The two instances illustrated in frame \textbf{b} of Figure \ref{fig:ro} occurred during the late growth stage traversal of row 4, which is one of the most challenging crop rows out of all the 10 crop rows. Both of these occurrences were caused by having longer fragment discontinuities in the crop row due to missing plants. Albeit our crop row detection system is able to recover the missing segments of the crop row~\citep{de2023deep}, it fails when there are longer segments of discontinuities in the crop row. There were no visible plants belonging to the crop row within the image frame during these two instances causing the crop row detection algorithm to falsely detect the adjacent row as the desired traversable crop row. The scenario \textbf{c} of Figure \ref{fig:ro} was also due to a similar perturbation where the early growth stage crop was not growing faster in certain parts of the crop row leading the crop row detection algorithm to attempt traversing to the adjacent crop row. The results presented in Sections \ref{sec:ne} and \ref{sec:fva} are calculated excluding these perturbations in the autonomous navigation trajectories. However, the effect of these instances in the data towards the overall accuracy figures presented was negligible.

\begin{figure}
\centering
\captionsetup{justification=raggedright,singlelinecheck=false}
\includegraphics[scale=0.26]{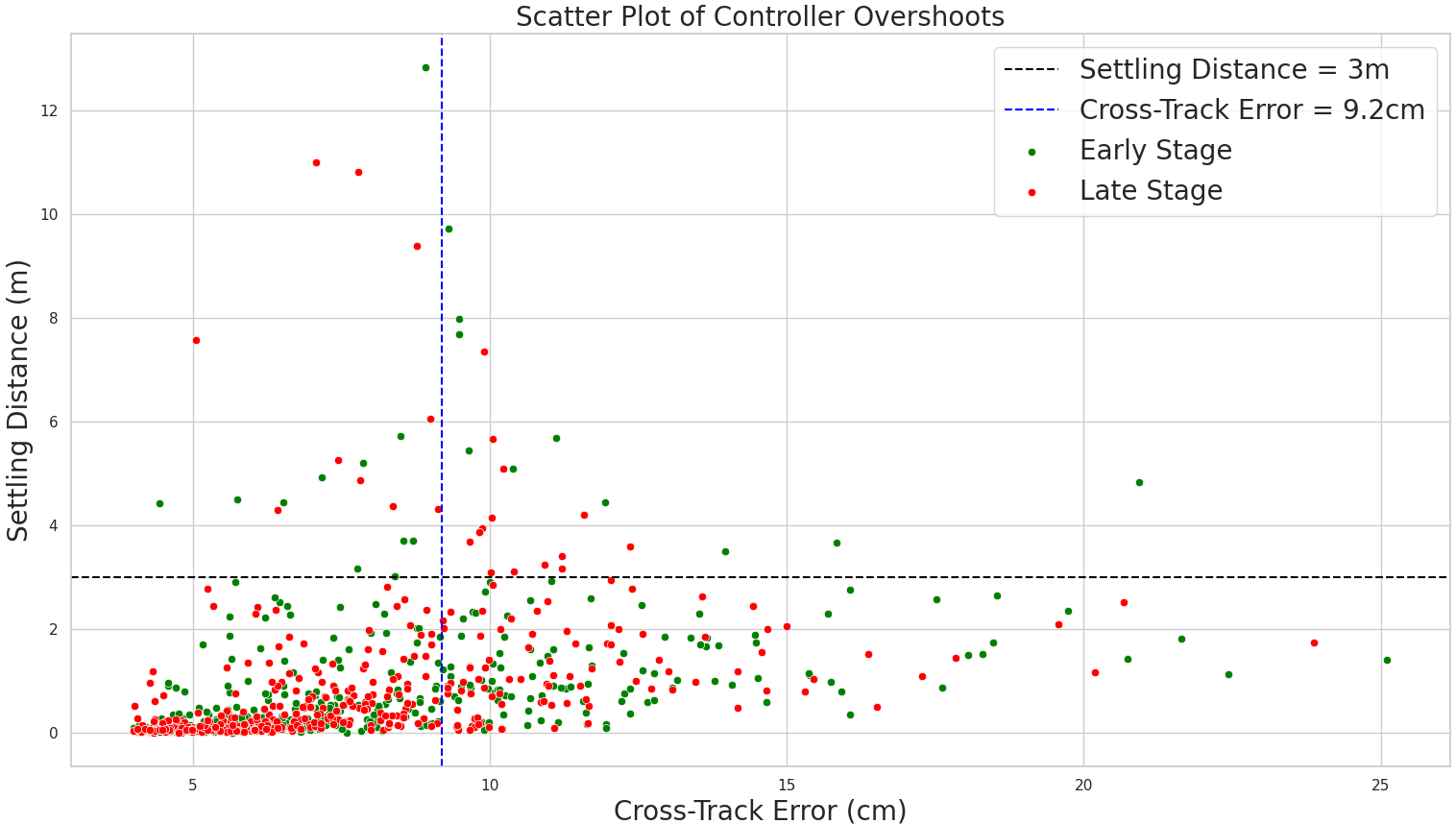}
\caption{Variation of settling distance vs. cross-track error during crop row following (Robot Speed = 0.3 ms\textsuperscript{-1})}
\label{fig:os}
\end{figure}

The oscillations in controller response during early growth stage navigation trials were observed to be relatively higher than that of the late growth stage navigation. However, the overall error during the early growth stage navigation was relatively lower compared to the late growth stage navigation trials. This observation suggests that the controller parameters could be optimized based on the growth stage for better response during autonomous navigation. The controller overshoots could be further observed by calculating the settling distance for above-average (> 4 cm) cross-track errors. The settling distances for above-average local maximums of the cross-track error curve during early-stage and late-stage crop row following trials were plotted in Figure \ref{fig:os}. The y-axis of this plot indicates a settling distance rather than a settling time due to the constant velocity of the robot set at 0.3 ms\textsuperscript{-1}. The scatter plot confirms that the majority of the above-average cross-track errors are recovered to a below-average offset within 3 m (or 10 seconds) of traversal forwards into the crop row. This also shows that the settling distance variation exhibits a peak around 9.22 cm cross-track error region with a standard deviation of 1.93 cm for cross-track errors with larger settling distance (>3 m). About 76\% of the data points belonging to this sub-distribution (mean=9.22 cm, stdev=1.93 cm) originate from 3 crop rows: 1,4 and 10. While sufficient evidence indicates that the controller can quickly respond to larger cross-track errors,  76\% of the data points belonging to this sub-distribution representing settling distances over 3 m (mean=9.22 cm, stdev=1.93 cm) were observed to be originating from 3 crop rows: 1,4 and 10. All these 3 crop rows are located closer to the edge of the field\footnote{Row 10 is located next to an agroforestry tree row, which could be considered as an edge of the field.}, where the crop rows are susceptible to having longer segments of missing plants due to foraging wildlife on the crop. 

\subsection{Experiment 2: Crop Row Switching}
\label{sec:ex2}

The row switching manoeuvre presented in Section \ref{sec:rsw} is a three-step process that involves 6 state transitions as illustrated in Figure \ref{fig:fsm}. The accuracy and the performance of each transition are individually evaluated during the experiments. The Hexman Mark-1 robot was programmed to follow a crop row and automatically detect the EOR and re-entry positions. The proposed row-switching manoeuvre is automatically started based on the EOR detection trigger. The path of the robot during the row-switching manoeuvre was tracked using the onboard RTK GNSS tracker with sub-centimetre accuracy. 

\begin{figure}
\centering
\captionsetup{justification=centering}
\includegraphics[scale=0.25]{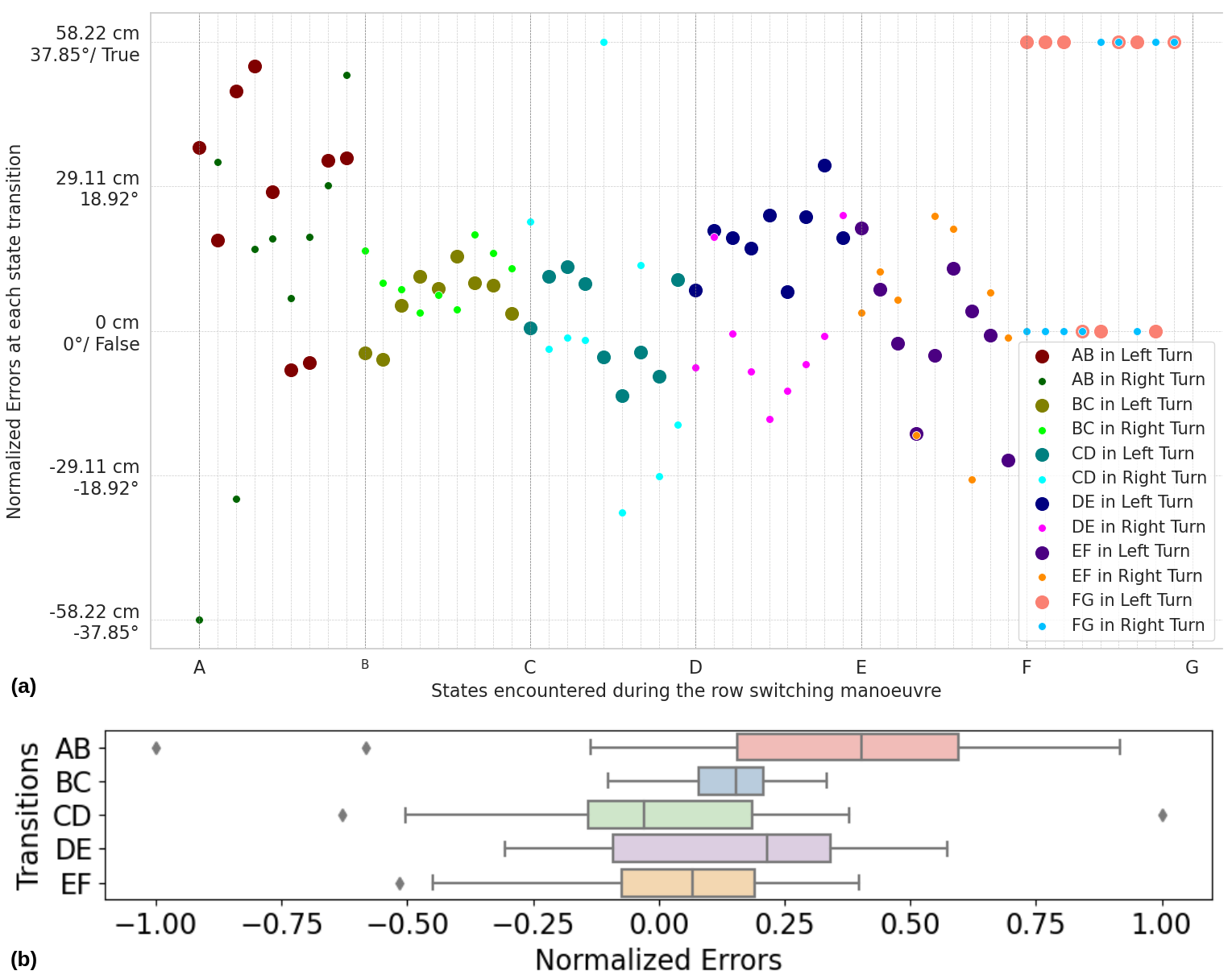}
\caption{Normalized state transition errors during the row switching manoeuvre. a: Scatter plot, b: Box and whisker plot.}
\label{fig:err}
\end{figure}

\begin{figure}
\centering
\captionsetup{justification=centering}
\includegraphics[scale=0.25]{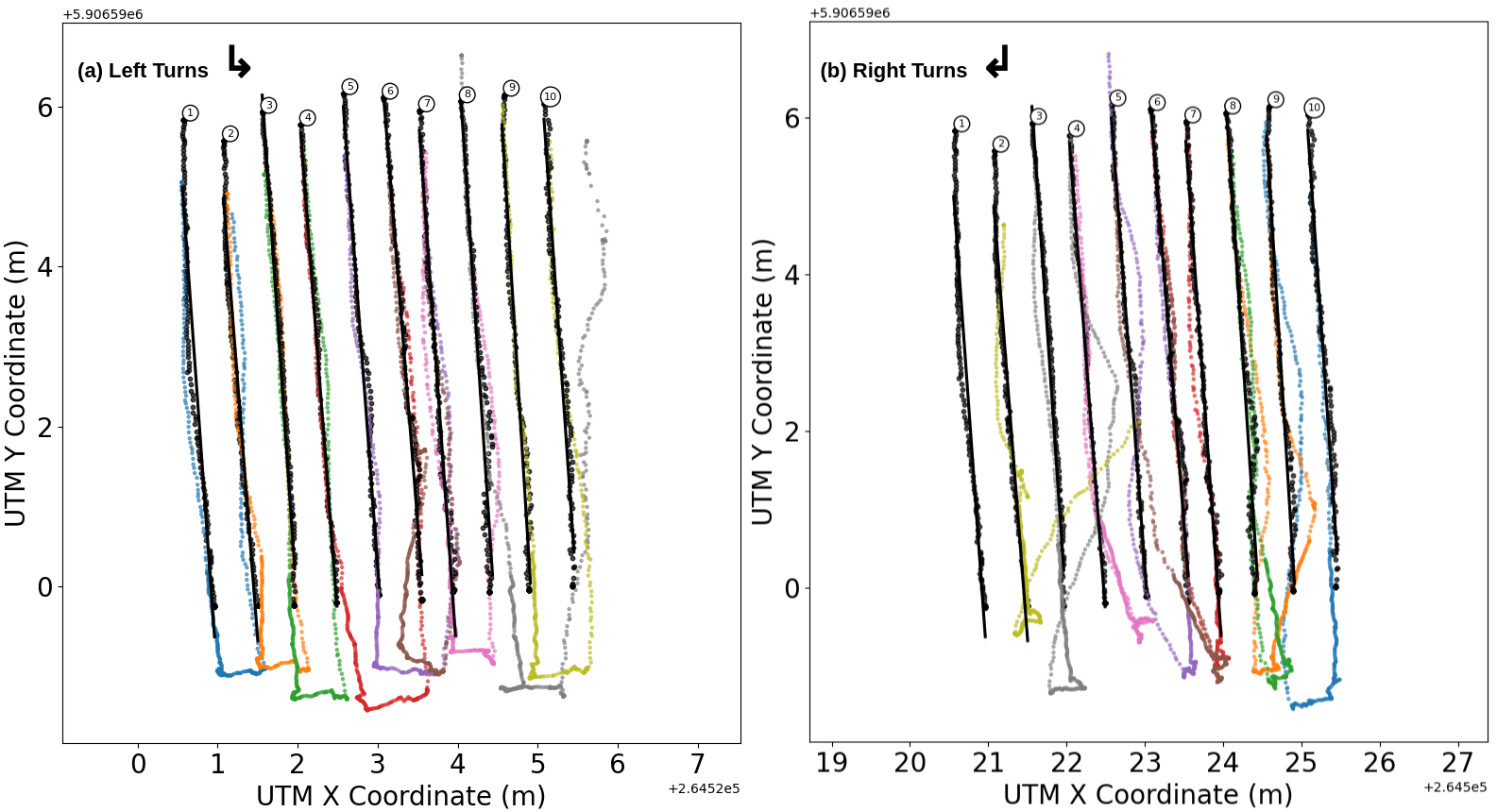}
\caption{UTM projections of the GNSS trajectories from row switching experiments (Black: Regression lines and ground truth coordinates).}
\label{fig:gpss}
\end{figure}

An experiment was set up in a selected area of 10 crop rows within a real sugar beet field. The GNSS coordinates of the 10 crop rows were recorded with sub-centimetre accuracy by driving the robot through each crop row at very slow speeds by an expert human driver. These ground truth GNSS coordinates of each crop row were then used to generate a regression line which would be used as a reference to calculate errors in autonomous navigation during row switching manoeuvre of the robot. The robot was allowed to autonomously execute the row switching manoeuvre among the 10 crop rows turning in both directions (left and right turns). A total of 18 row-switching trials were conducted during this experiment as plotted in Figure \ref{fig:gpss}. The GNSS coordinates were converted to Universal Transverse Mercator (UTM) coordinates for plotting and error calculations. The errors during each state transition of each row switching manoeuvre trial are illustrated in Figure \ref{fig:err}.a where distance errors and angular errors are normalized within a scatter plot. There are 9 x-ticks in Figure \ref{fig:err}.a between each pair of adjacent states which represents a pair of consecutive real-world crop rows where the trial took place. A box and whisker plot of the errors for each transition is denoted in Figure \ref{fig:err}.b. This normalised representation provides a comparative illustration of the error magnitudes in each state transition. The plots corresponding to translational steps stand above the zero line while box and whisker plots of angular transitions are centered close to zero. Table \ref{tab:err} presents the median errors of traversal during each stage of the row-switching manoeuvre. Normalized median percentage error ($\alpha = E_{median}/E_{max,T}$) was calculated where $E_{max,T}$ is the maximum absolute error for the type $T$ (T being a distance or an angle) across the entire manoeuvre and $E_{median}$ is the median error for each transition. The vision-based transition $A\rightarrow B$ records the highest $\alpha$ value representing the transition with the highest error. The remaining transitions record relatively lower $\alpha$ values representing accurate navigation relative to the vision-based transition.

\begin{table}
\centering
\caption{Linear and angular errors during each transition of the row switching maneuver.}
\begin{tabular}{|c|c|c|c|}
\hline
\multirow{2}{*}{Transition} & \multicolumn{2}{c|}{Median Errors} & \multirow{2}{*}{$\alpha$} \\

& Error & Absolute Error & \\
\hline
$A\rightarrow B$ & 23.40 cm & 31.63 cm & 40.20 \%\\
\hline
$B\rightarrow C$ & 8.87 cm & 8.87 cm & 15.24 \% \\
\hline
$C\rightarrow D$ & -1.09$^\circ$ & 6.91$^\circ$ & -2.88 \% \\
\hline
$D\rightarrow E$ & 12.47 cm & 17.26 cm & 21.41 \%\\
\hline
$E\rightarrow F$ & 2.51$^\circ$ & 6.62$^\circ$ & 6.64 \% \\
\hline
\end{tabular}
\label{tab:err}
\end{table}

\begin{figure}
\centering
\captionsetup{justification=centering}
\includegraphics[scale=0.45]{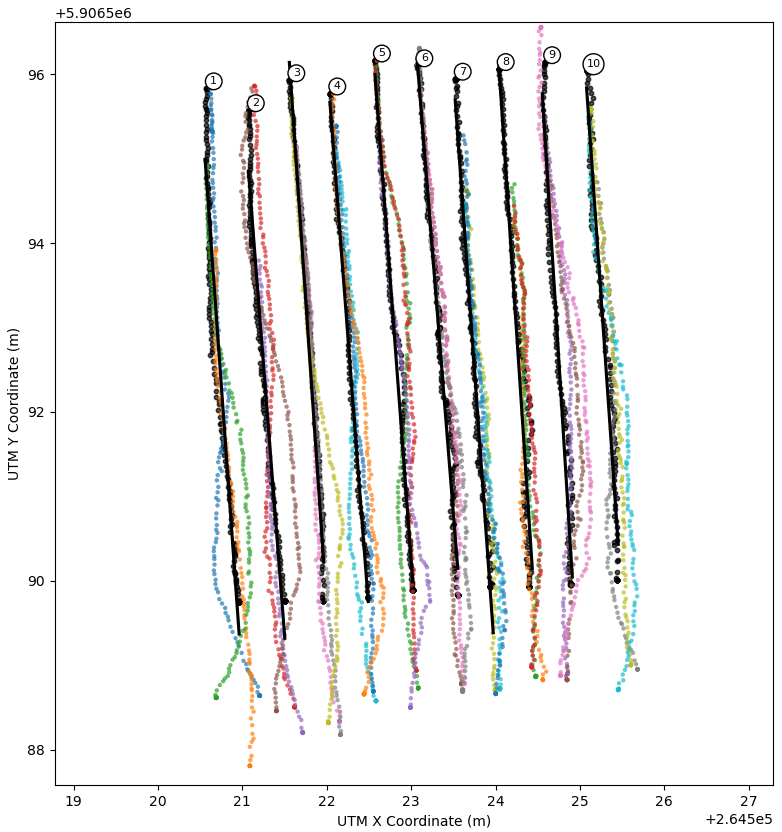}
\caption{GNSS trajectories from re-entry experiment (Black: Ground truth coordinates).}
\label{fig:retr}
\end{figure}

\subsubsection{Crop row exit and headland entry}
The row exit step ($A\rightarrow B\rightarrow C$ ) in the proposed row switching manoeuvre could be identified with the densely distributed GNSS coordinates in Figure \ref{fig:gpss} leading towards headland from within the crop row in each trial.  The dense distribution of GNSS coordinates attributed to the slower speed of row switching manoeuvre relative to the in-row navigation speed of the crop row navigation algorithm. 

The row crop row exit ($A\rightarrow B$ transition) is a vision-based navigation stage where the headland entry ($B\rightarrow C$ transition) uses the wheel odometry to guide the robot into the headland area. This difference in feedback modalities is reflected in the distance errors of each transition. The visual feedback in $A\rightarrow B$ transition ensures that the robot has successfully reached the EOR position with visual confirmation. Therefore, it is vital for successfully reaching the EOR despite the higher error margins compared to wheel odometry. The majority of the errors in $A\rightarrow B$ and $B\rightarrow C$ transitions are positive, which indicates that the robot always travels further into the headland area beyond desired positions at states $B$ and $C$. This trend does not have significant adverse impact on the overall row-switching manoeuvre since such extra distance traversed into the headland wouldn't cause the robot to damage crops during the U-turn step.

The robot would stop at 52.6 cm ($L_{robot}$) away from the actual EOR position in an ideal $C$ state. However, the overall maximum error in the row exit step ($E_{ABC,max}$) was recorded as 64.27 cm, a distance at which the robot would move further away from the desired position at state C. Based on these observations, the minimum width $W_{H,min}$ for the headland space was calculated to be 143.17 cm using Equation \ref{eq:whm}. The coefficient of $L_{robot}$ term in Equation \ref{eq:whm} was set to $1.85$ since the RTK-GNSS receiver used to measure the robot motion was mounted at 45 cm behind the front of the robot. This coefficient of $L_{robot}$ term must be changed to $(1+R)$ where $R$ depends on the position of the RTK-GNSS receiver on the robot if this experiment is being repeated on a different robot. The robot was expected to be aligned with the crop row at state $A$ within a heading error margin of 2$^\circ$. A heading error beyond 2$^\circ$ at state $A$ leads the robot to cross into the headland buffer of an adjacent crop row at state $C$, which would cause it to skip 1 crop row during switching or re-enter the same crop row it traversed as seen in the unsuccessful attempts in Figure \ref{fig:gpss}. 

\begin{equation} \label{eq:whm}
  W_{H,min} = 1.85\times L_{robot} + E_{ABC,max}
\end{equation}

\subsubsection{U-turn towards next crop row}
The U-turn step of the row-switching manoeuvre is represented by the state transitions $C\rightarrow D\rightarrow E\rightarrow F$. The angular error for $C\rightarrow D$ transition was calculated based on the angle between $\overrightarrow{AC}$ and $\overrightarrow{DE}$ vectors. The angle between $\overrightarrow{DE}$ and $\overrightarrow{FF_{N}}$ vectors was considered to calculate the angular error for $E\rightarrow F$ transition where $F_{N}$ is a point on the GNSS trajectory which is $N$(=5) points after the GNSS coordinate of state F. Distance error for $D\rightarrow E$ transition was calculated by comparing the $DE$ distance with the inter-row distance between the adjacent crop rows which the robot is being switched, using regressed ground truth lines.

The angular errors in both rotational transitions are evenly distributed with a near-zero mean. The absolute median angular errors during rotational transitions were less than 7$^\circ$. These error margins could be considered acceptable for this application scenario since such small angular errors wouldn't incur significant deviations to $\overrightarrow{DE}$ and $\overrightarrow{FG}$ vectors. The robot would stay at the same place without any translational motion in an ideal rotational transition. However, it was evident from some of the recorded trajectories in Figure \ref{fig:gpss} that these rotations also incur some translational motion within the trajectory. Such motions push the robot towards or away from the direction of the next crop row. This would cause the robot to skip the next crop row to be traversed or turn towards the same row it came in, despite achieving the desired $\overrightarrow{DE}$ distance. This unintended translational motion is often caused by the uneven terrain in the headland area which is not detected by the wheel odometry. The headland buffer area of $4^{th}$ crop row had such uneven terrain which led to the failure of both left and right turns originating from that crop row.

\subsubsection{TSM-based Re-Entry Validation}
\label{sec:rev}
The median error for  $D\rightarrow E$ transition which led to successful re-entry was always below 30 cm. This indicates that the robot could execute a successful re-entry when it faces the next crop row with a perpendicular offset of 30 cm or below at state $F$. An experiment was set up to validate this hypothesis where the robot was placed facing towards the crop row at different angles within the headland buffer of a given crop row. The TSM algorithm was executed on the robot such that it would detect the crop row in front of it and gradually drive the robot into the crop row in front. The path of the robot was recorded in GNSS coordinates as plotted in Figure \ref{fig:retr}. All the recorded trajectories could successfully enter the row in front of it since the robot was initiated in the headland buffer and facing towards the general direction of the crop row in front of it. 

The re-entry failures in Figure \ref{fig:gpss} could be explained by the main findings of this experiment. There are two key factors governing the success of re-entry to the next crop row during $F\rightarrow G$ transition. The first requirement is that the robot must be positioned within the headland buffer of the crop row it intends to enter. If the perpendicular offset between the crop row and robot position at state $F$ is beyond 30 cm, a re-entry failure occurs. The second factor is that the robot must be oriented towards the general direction of the crop row it intends to enter. The maximum deviation angle of the robot heading from crop row was 26$^\circ$ in the experiment illustrated in Figure \ref{fig:retr}. Although the errors in individual state transitions of the row switching manoeuvre are minimal, the overall outcome of all transitions would not lead to success when these two requirements are not met at state $F$. 

\subsection{Experiment 3: Field Scale Navigation}
\label{sec:ex3}

\begin{figure}
\centering
\captionsetup{justification=raggedright,singlelinecheck=false}
\includegraphics[scale=0.3]{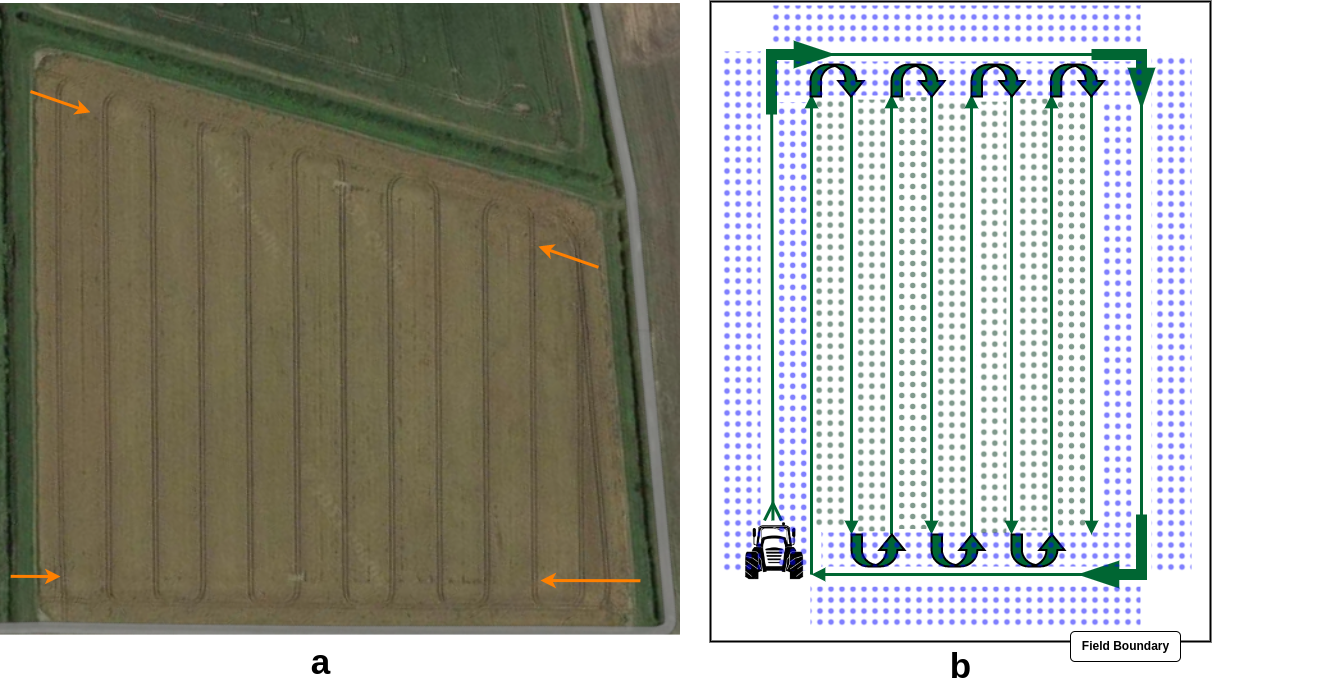}
\caption{Field structure of a real arable field. a: A satellite image of a real arable field (Copyright: Google Maps), b: The crop structure in a real arable field}
\label{fig:fld}
\end{figure}

Figure \ref{fig:fld} demonstrates a side-by-side illustration of the crop structure of a real arable field. The tractor initially enters the field and starts drilling around the perimeter as illustrated in Figure \ref{fig:fld}.b. Then it fills in the center of the field in a lawnmower pattern. The plants drilled around the perimeter are indicated in blue in Figure \ref{fig:fld}.b while the plants in the centre are marked in green. The demarcation line between non-parallel blue and green plant regions is indicated with orange arrows in Figure \ref{fig:fld}.a. The crop row switching behaviour proposed in this work is only applicable in the regions of the field indicated in blue. Attempting to execute the crop row switching behaviour in the green region of a real arable field would cause the robot to drive over the crops in the blue region. A field-scale deployment of this proposed infield navigation scheme in a real-world arable field structured for optimal navigation of tractors is limited due to the field structure explained above. Therefore, the field coverage analysis experiment was conducted in a simulated field where the entire field constitutes parallel crop rows with headland space on either side of the ends of the crop rows.

\subsubsection{initial turning direction detection Analysis}
\begin{figure}
\centering
\captionsetup{justification=raggedright,singlelinecheck=false}
\includegraphics[scale=0.49]{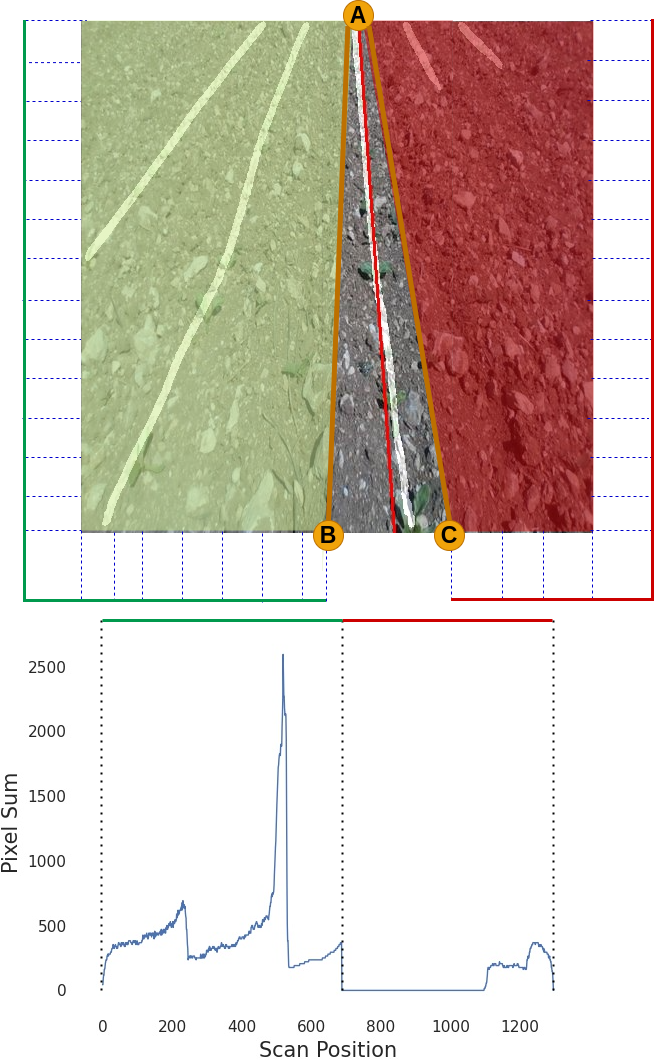}
\caption{Field structure of a real arable field. a: A satellite image of a real arable field (Copyright: Google Maps), b: The crop structure in a real arable field.}
\label{fig:fp}
\end{figure}

The initial turning direction detection algorithm described in Section \ref{sec:vbifo} was tested with the test dataset of the CRDLD dataset. There were 500 images in the test dataset and 10 of them were images of crop rows next to the field edge. The algorithm was able to successfully detect 100\% of the images next to the field edge with the correct initial turn direction. However, one false positive was detected during the testing as illustrated in Figure~\ref{fig:fp}. This image was captured in a crop row in the middle of the field. The predicted crop row mask only illustrated the top portions of the crop rows detected on the right side of the central crop row leading the algorithm to predict a false positive edge row.

\subsubsection{Field Coverage Analysis}

\begin{figure}
\centering
\captionsetup{justification=raggedright,singlelinecheck=false}
\includegraphics[scale=0.29]{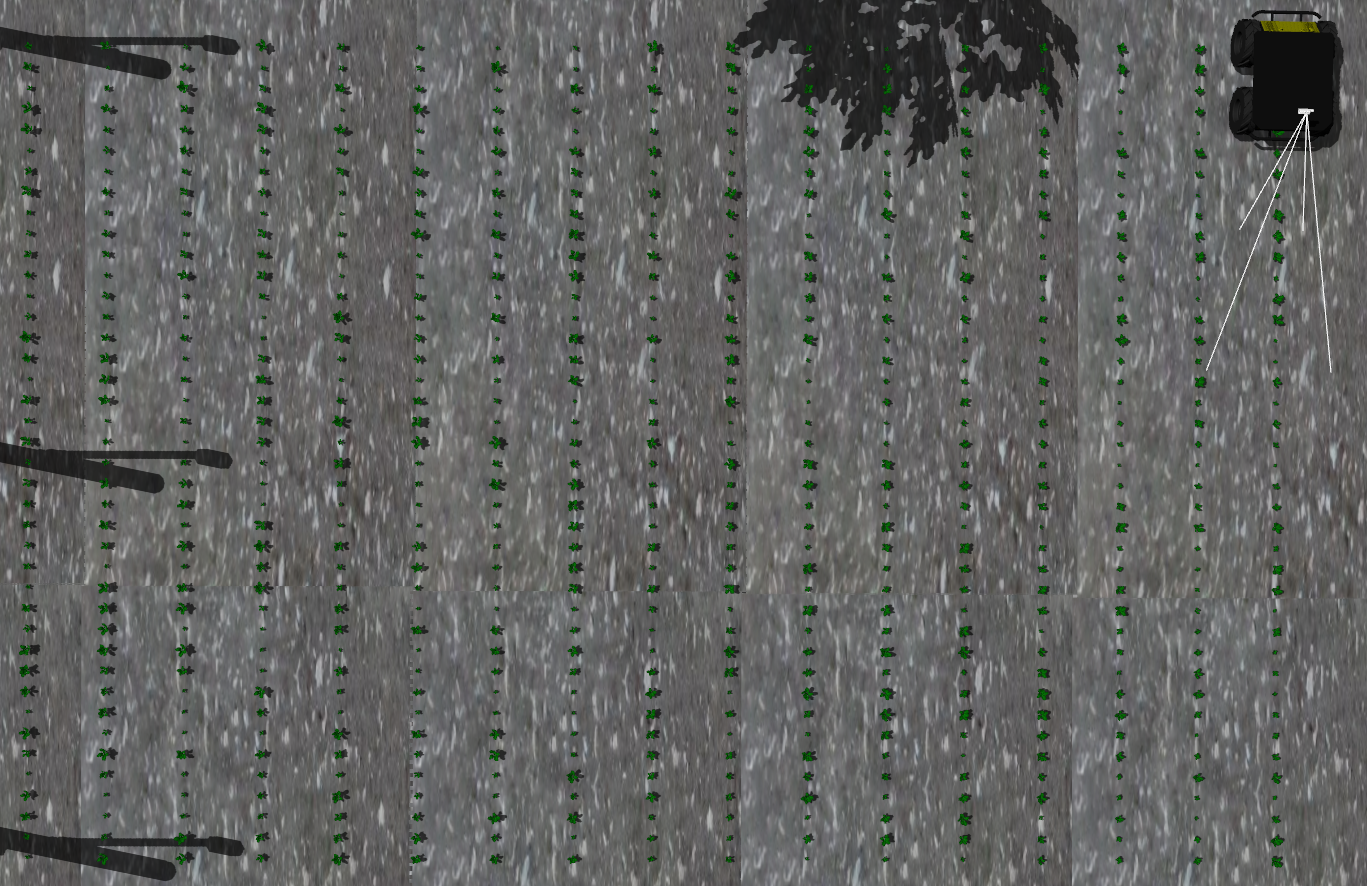}
\caption{Simulated sugar beet field with the Husky robot at the starting position of the crop row.}
\label{fig:sim}
\end{figure}

The aim of this experiment is to explore the ability of the proposed scheme to navigate an entire arable field while alternating between row-following and row-switching behaviours. To this end, a simulation environment was set up in the Gazebo simulator with sugar beet plants with realistic textures and the ground texture of the field was transferred from the real environment to the simulation. The parameters pertaining to this simulation environment are listed in Table \ref{tab:smp}. Plant height and orientation were randomized within the stated variances to mimic realistic variations among the plants in the field as shown in Figure \ref{fig:sim}. 

\begin{table}
\centering
\caption{Simulation Parameter for Sugar Beet Field}
\begin{center}
\begin{tabular}{|p{0.25\linewidth} | p{0.12\linewidth} | p{0.14\linewidth}| p{0.17\linewidth}|}
\hline
\textbf{Property} & \textbf{Value} & \textbf{Variance}\\
\hline
{Row Spacing} & 60 cm & 0 \\ 
\hline
{Seed Spacing} & 16 cm & 0 \\ 
\hline
{Plant Height} & 6 cm & +3 cm \\ 
\hline
{Plant Orientation} & 0$^\circ$ & $\pm$ 145$^\circ$ \\ 
\hline
Row Length & 6 m & 0 \\ 
\hline
{Row Count} & 20 & 0 \\ 
\hline
\end{tabular}
\label{tab:smp}
\end{center}
\end{table}

This experiment validates initial turning direction detection algorithm introduced in Section \ref{sec:vbifo} and the impact of row following and row switching behaviour transitions on field coverage of the proposed infield navigation scheme. The robot will first initiate its infield orientation algorithm to identify the turning direction during its first crop row-switching manoeuvre and continue with row-following behaviour. It will start the row-switching behaviour when it detects its approach towards the end of the row and continues with alternating behaviours until the last crop row. It will stop the navigation scheme when it reaches the last crop row based on the initial turning direction detection algorithm. The robot was positioned at each corner of the simulated sugar beet field similar to its position in Figure \ref{fig:sim} and the autonomous navigation scheme was initiated. 

\begin{figure}
\centering
\captionsetup{justification=raggedright,singlelinecheck=false}
\includegraphics[scale=0.22]{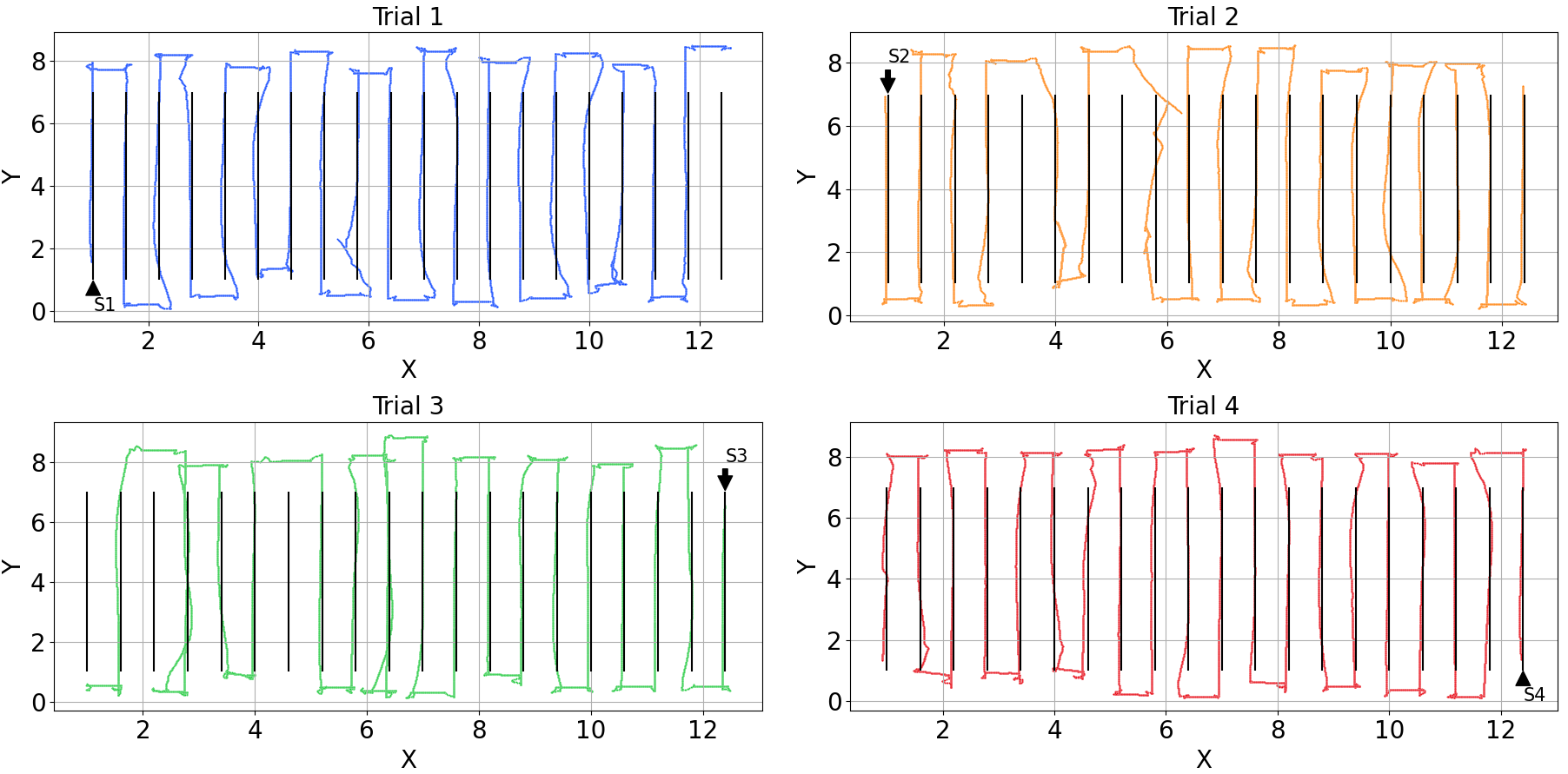}
\caption{Autonomous field scale navigation trajectories for simulated sugar beet field navigation trials. (Each trial starts from each corner of the field indicated by starting points S1-S4).}
\label{fig:smt}
\end{figure}

The autonomous navigation trajectories of the robot during each of the four trials are visualised in Figure \ref{fig:smt}. The robot has traversed 80 crop rows in total where it failed to traverse 6 crop rows. Therefore the percentage field coverage of all the trials was 92.5\%. There were 2 instances in trial 3 in which the robot re-enters back into the same crop row it traversed. Therefore the traversal overlap percentage in the experiment was 2.5\%. These errors originated during the $E\rightarrow F$ transition of the crop row switching manoeuvre illustrated in Figure \ref{fig:utr}. The raw skipping occurs when the robot underturns during $E\rightarrow F$ transition causing the robot to orient itself to the adjacent crop row rather than the intended crop row as seen in row skipping instances of trial 3 of Figure \ref{fig:smt}. This causes the robot to traverse further than the intended inter-row distance to reach the next crop row to be traversed. The traversal overlap events occur when the robot overturns the expected 90$^\circ$ angle during $E\rightarrow F$ transition, turning back towards the next crop row to be traversed. The row-following algorithm latches the robot towards the already traversed crop row since it is oriented towards the same crop row it already traversed. Both of the described anomalies could be corrected by improving the turning behaviour of the robot turning stage. This could be achieved by employing a predictive controller to avoid overshoots and undershoots, and using sensor fusion for additional positional validation. However, these overshoots and undershoots were not prominent during the turning stages of the real-world crop row-switching experiments conducted in this work. The delay in updating the simulation due to limited computing resources of the system on which the simulator was run could explain the under-performance observed during the simulation. There were two instances where the human driver manually intervened to re-align the robot at the 9\textsuperscript{th} crop row from left during trials 1 and 2.

\section{Conclusions and Future Work}
\label{sec:con}

A vision-based arable field navigation scheme is presented in this paper. The proposed approach combines alternating robot behaviours for crop row following and crop row switching to achieve field scale navigation. A initial turning direction detection algorithm was proposed to autonomously determine the initial turning direction for a robot starting the field traversal from any corner of the field without pre-assigned directions. The row following behaviour was tested extensively under varying field conditions and growth stages. The autonomous row following behaviour of the robot was tested during a 4.5 km infield navigation experiment which yielded an average of 3.32 cm median cross-track error and 1.24$^\circ$ median heading error. The row following algorithm was challenged due to larger (>1 m) discontinuities in the crop rows where human intervention was needed to re-orient the robot to the crop row. 

The proposed row-switching manoeuvre could navigate the robot from one crop row to another without needing RTK-GNSS sensors or multiple cameras while using a single front-mounted camera on the robot. Individual steps of the row-switching manoeuvre demonstrated excellent results within the context of each state transition and its functionality. The vision-based $A\rightarrow B$ transition was the transition with the highest errors in the proposed manoeuvre. The rotational transitions of the manoeuvre yielded smaller percentage errors relative to the translational transitions.  The success rate of the conducted real world row-switching experiment was 55.5\% while the re-entry experiment yielded a 100\% success rate. In contrast, the success rate of the simulated row switching experiment (Section \ref{sec:ex3}) was recorded at 89.5\%.The row-switching manoeuvre is also crop agnostic since it doesn't rely on plant-specific visual features. 

The row following behaviour was combined with the vision-based crop row switching algorithm in a simulated field scale navigation experiment recording a 92.5\% field coverage and 2.5\% crop row traversal overlap. The row skipping and traversal overlap errors were mainly caused due to the inaccurate rotational movements during the crop row switching manoeuvre. The initial turning direction detection algorithm was tested on the CRDLD dataset with a 100\% success rate and re-validated in the simulated deployment of field scale navigation. The difference in success rates of row switching manoeuvre in simulation and the real world could be accounted for by the uneven terrains in the real world and the lack of IMU fused odometry in the robot used in the real-world experiment. 

The larger discontinuities are a hard barrier for the current system as the crop row detection and row following commands are generated frame by frame to navigate the robot along the crop row. While complementary filters are implemented to filter out sudden noise in the detected crop row, those filters are not sufficient to prevent the robot from wandering off the traversing crop row path when large discontinuities are encountered. The main problem in overcoming the challenge of larger discontinuities is that the system is unable to distinguish between a large gap in the crop row and the end of the crop row. The information provided by the vision system in both scenarios is largely similar. A software-based solution for such large crop row gaps is to generate a virtual navigation line based on the presence of adjacent crop rows. However, this solution is not very promising in practical settings mainly because large gaps in crop rows are associated with missing crops due to tramlines or foraging wildlife. The likelihood of having lateral crop rows intact in such scenarios is very low. A hardware solution could be suggested with higher confidence to address this problem where the camera could be tilted upwards to capture a larger section of succeeding crop row within the image frame. 

Two main unexpected behaviour patterns in the row exit and U-turn steps lead to the failure of row switching: large heading error at state $A$ and translational motion during rotational state transitions. These two shortcomings of the proposed row switching manoeuvre could be corrected by introducing a heading correction step at state $A$ and inertial measurement unit (IMU) based sensor fusion to track the unintended translational motion during the rotational transitions correcting the intended $\overrightarrow{DE}$ distance. 

The vision-based infield navigation scheme proposed in this thesis was able to achieve 92\% field coverage and 2.5\% traversal overlap as reported in Chapter 7. However, further development would be necessary in order to reach the ideal 100\% field coverage and 0\% traversal overlap. While row-following behaviour delivered promising results, crop row-switching algorithms faced significant challenges navigating the headland area mainly using wheel odometry. The errors and challenges encountered during the headland traversal could be mitigated by incorporating additional modalities for localisation during the row-switching manoeuvre. This could be achieved by incorporating visual SLAM algorithms and IMU data during the crop row switching as sensor fusion with the wheel odometry.

\bibliography{sample}
\end{document}